\documentclass[10pt,twocolumn,letterpaper]{article}

\usepackage{wacv}
\usepackage{times}
\usepackage{epsfig}
\usepackage{graphicx}
\usepackage{amsmath}
\usepackage{amssymb}
\usepackage{booktabs}
\usepackage{microtype}

\graphicspath{{./graphics/}}
\usepackage{animate}
\usepackage{tabularx}
\usepackage{booktabs}
\usepackage[labelsep=period,font=small]{caption}
\usepackage{tikz}
\usepackage{pgfplots}
\usepackage{pgfplotstable}
\usepackage[skins]{tcolorbox}
\usepackage{standalone}
\usepackage{bm}
\usepackage{mleftright}
\usepackage{nicefrac}
\usepackage{makecell}
\usepackage{textcomp}
\usepackage[normalem]{ulem}
\usepackage[shortcuts]{extdash}
\usepackage[outline]{contour}
\newlength{\itemwidth}

\newcolumntype{Y}{>{\centering\arraybackslash}X}
\newcolumntype{P}[1]{>{\centering\arraybackslash}p{#1}}

\usetikzlibrary{calc}
\usetikzlibrary{tikzmark}
\usetikzlibrary{spy}
\usetikzlibrary{shapes.misc}
\usetikzlibrary{fadings}
\pgfplotsset{compat=newest}

\definecolor{EE7F0E}{RGB}{238,127,14}
\definecolor{3787CF}{RGB}{55,135,207}
\definecolor{619D47}{RGB}{97,157,71}
\definecolor{DBDC4A}{RGB}{219,220,74}
\definecolor{F1C232}{RGB}{241,194,50}

\pgfplotscreateplotcyclelist{custompalette}{
    {EE7F0E!90!black,fill=EE7F0E},
    {3787CF!90!black,fill=3787CF},
    {619D47!90!black,fill=619D47},
    {DBDC4A!90!black,fill=DBDC4A}
}
\pgfplotscreateplotcyclelist{anothercustompalette}{
    {EE7F0E!90!black,line width=1.5pt},
    {3787CF!90!black,line width=1.5pt},
    {619D47!90!black,line width=1.5pt},
    {DBDC4A!90!black,line width=1.5pt}
}
\pgfplotscreateplotcyclelist{anothercustompalette_rev}{
    {619D47!90!black,line width=1.5pt},
    {3787CF!90!black,line width=1.5pt},
    {EE7F0E!90!black,line width=1.5pt}
}

\usepackage{stmaryrd}
\usepackage{trimclip}

\makeatletter
\DeclareRobustCommand{\shortto}{%
  \mathrel{\mathpalette\short@to\relax}%
}

\DeclareRobustCommand{\veryshortto}{%
  \mathrel{\mathpalette\veryshort@to\relax}%
}

\newcommand{\short@to}[2]{%
  \mkern2mu
  \clipbox{{.3\width} 0 0 0}{$\m@th#1\vphantom{+}{\shortrightarrow}$}%
  }

\newcommand{\veryshort@to}[2]{%
  \mkern2mu
  \clipbox{{.2\width} 0 0 0}{$\m@th#1\vphantom{+}{\shortrightarrow}$}%
  }
\makeatother

\newcommand{\usolid}[1]{%
    \tikz[remember picture, baseline=(tosolid.base)]{
        \node[inner sep=0pt, outer sep=0pt] (tosolid) {#1};
    }%
    \tikz[remember picture, overlay]{
        \draw[] ([yshift=-1.5pt]tosolid.south west) -- ([yshift=-1.5pt]tosolid.south east);
    }%
}%

\newcommand{\udotted}[1]{%
    \tikz[remember picture, baseline=(todotted.base)]{
        \node[inner sep=0pt, outer sep=0pt] (todotted) {#1};
    }%
    \tikz[remember picture, overlay]{
        \draw[densely dotted, line width=0.5] ([yshift=-1.5pt]todotted.south west) -- ([yshift=-1.5pt]todotted.south east);
    }%
}%

\wacvalgorithmstrack
\wacvfinalcopy

\usepackage[breaklinks=true,bookmarks=false]{hyperref}

\begin{document}

%%%%%%%%% TITLE
\title{Splatting-based Synthesis for Video Frame Interpolation}

\makeatletter
\g@addto@macro\@maketitle{
    \vspace*{-13pt}
    \begin{minipage}[t]{0.478\textwidth}
        \par\vspace{0.0cm}
        \hspace{-0.1cm}\includegraphics[]{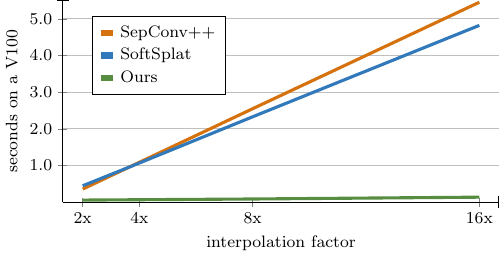}\vspace{-0.2cm}
        \captionof{figure}{Runtime of two common video frame interpolation approaches versus ours when interpolating multiple frames between two inputs from XTEST-2K~\cite{Sim_ICCV_2021}. Our proposed approach interpolates the first frame in $61$ ms and each additional frame only takes a few milliseconds thanks to our splatting-based synthesis.}\vspace{-0.0cm}
        \label{fig:multime}
    \end{minipage}\hfill\begin{minipage}[t]{0.478\textwidth}
        \par\vspace{0.0cm}
        \hspace{-0.1cm}\includegraphics[]{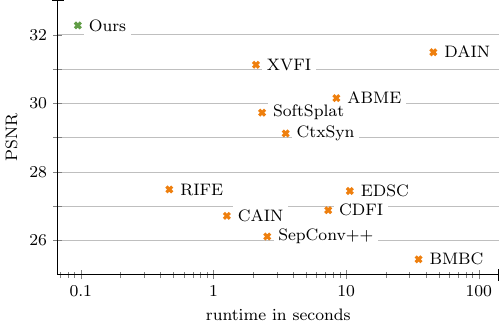}\vspace{-0.2cm}
        \captionof{figure}{Evaluating the $8 \times$ interpolation ability of our proposed approach in comparison to various others on XTEST-2K~\cite{Sim_ICCV_2021}.}\vspace{-0.0cm}
        \label{fig:perfplot}
    \end{minipage}
    \vspace*{20pt}
}
\makeatother

\author{
Simon Niklaus\\
{\small Adobe Research\vspace{0.8cm}}
\and
Ping Hu\\
{\small Boston University\vspace{0.8cm}}
\and
Jiawen Chen\\
{\small Adobe Inc\vspace{0.8cm}}
}

\maketitle

\thispagestyle{empty}

\begin{abstract}

Frame interpolation is an essential video processing technique that adjusts the temporal resolution of an image sequence. While deep learning has brought great improvements to the area of video frame interpolation, techniques that make use of neural networks can typically not easily be deployed in practical applications like a video editor since they are either computationally too demanding or fail at high resolutions. In contrast, we propose a deep learning approach that solely relies on splatting to synthesize interpolated frames. This splatting-based synthesis for video frame interpolation is not only much faster than similar approaches, especially for multi-frame interpolation, but can also yield new state-of-the-art results at high resolutions.

\end{abstract}

\vspace*{-0.3cm}
\section{Introduction}
\label{sec:intro}

Video frame interpolation is becoming more and more ubiquitous. While early techniques for frame interpolation were restricted to using block motion estimation and compensation due to performance constraints~\cite{Choi_TCE_2000, Ha_TCE_2004}, modern graphics accelerators allow for dense motion estimation and compensation while heavily making use of neural networks~\cite{Liu_ICCV_2017, Niklaus_CVPR_2017, Niklaus_ICCV_2017, Nikltwo_WACV_2021}. These developments enable interesting new applications of video frame interpolation for animation inbetweening~\cite{Sili_CVPR_2021}, video compression~\cite{Wu_ECCV_2018}, video editing~\cite{Meyer_BMVC_2018}, motion blur synthesis~\cite{Brooks_CVPR_2019}, and many others.

However, current interpolation techniques that make use of neural networks are inherently difficult to accelerate. For example, the first interpolation approaches that use deep learning require fully executing the entire network for each output~\cite{Liu_ICCV_2017, Niklaus_CVPR_2017, Niklaus_ICCV_2017}. As such, using SepConv++~\cite{Niklone_WACV_2021} (Figure~\ref{fig:multime}, orange) to interpolate a video by a factor of $8 \times$ instead of $2 \times$ requires eight times more compute. Newer approaches are little different though, SoftSplat~\cite{Niklaus_CVPR_2020} (Figure~\ref{fig:multime}, blue) for instance estimates the optical flow between the input frames and then extracts and warps feature pyramids to the desired instant before employing a synthesis network to yield the final result. While the optical flow only needs to be estimated once in this case, the synthesis network has to be executed for each new frame which again requires roughly eight times more compute when interpolating by $8 \times$ instead of $2 \times$.

\begin{figure*}\centering
    \setlength{\tabcolsep}{0.05cm}
    \setlength{\itemwidth}{4.27cm}
    \hspace*{-\tabcolsep}\begin{tabular}{cccc}
            \begin{tikzpicture}[spy using outlines={3787CF, magnification=4, width={\itemwidth - 0.06cm}, height=2.4cm, connect spies,
    every spy in node/.append style={line width=0.06cm}}]
                \node [inner sep=0.0cm] {\includegraphics[width=\itemwidth]{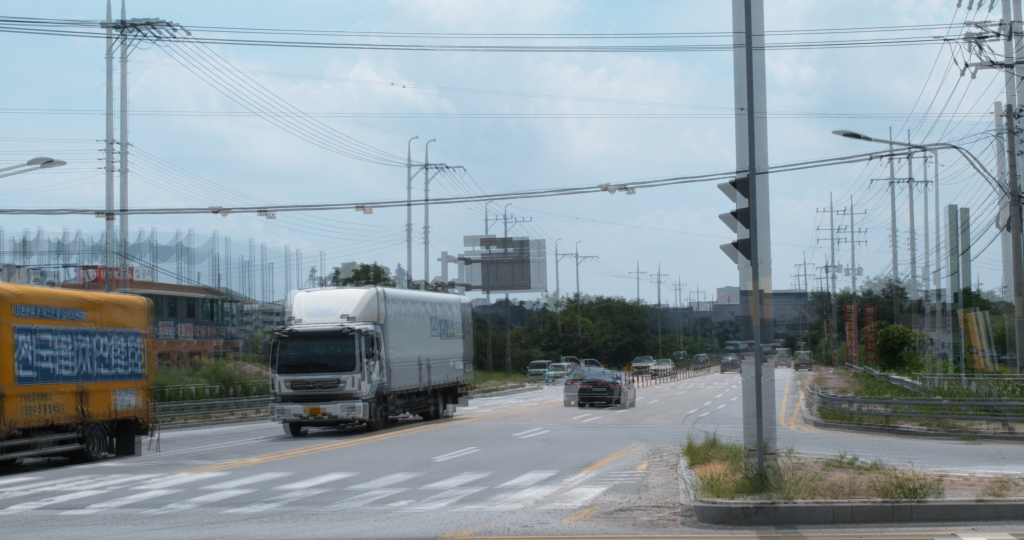}};
                \spy [every spy on node/.append style={line width=0.06cm}, spy connection path={\draw[line width=0.06cm] (tikzspyonnode) -- (tikzspyinnode);}] on (0.8,-0.27) in node at (0.0,-2.45);
            \end{tikzpicture}
        &
            \begin{tikzpicture}[spy using outlines={3787CF, magnification=4, width={\itemwidth - 0.06cm}, height=2.4cm, connect spies,
    every spy in node/.append style={line width=0.06cm}}]
                \node [inner sep=0.0cm] {\includegraphics[width=\itemwidth]{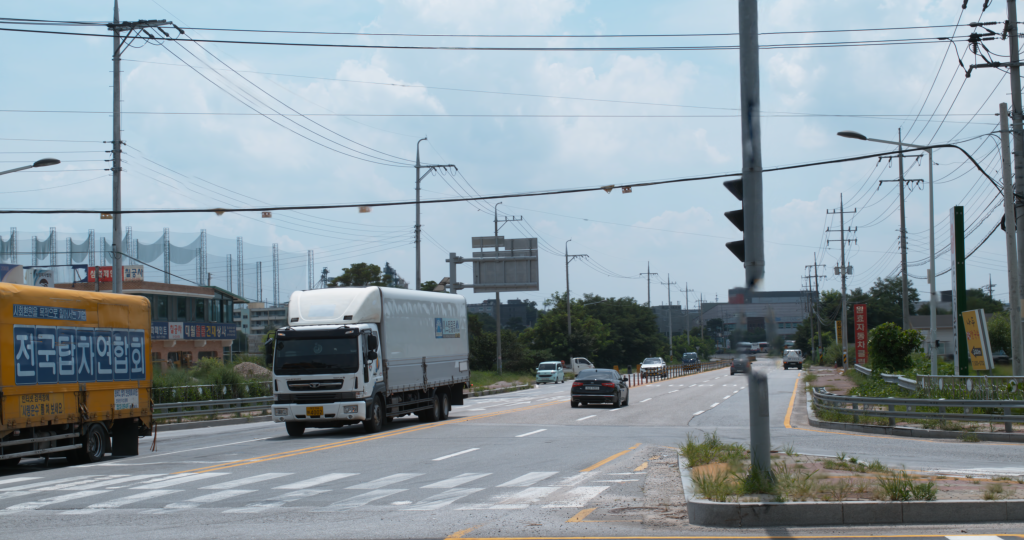}};
                \spy [every spy on node/.append style={line width=0.06cm}, spy connection path={\draw[line width=0.06cm] (tikzspyonnode) -- (tikzspyinnode);}] on (0.8,-0.27) in node at (0.0,-2.45);
            \end{tikzpicture}
        &
            \begin{tikzpicture}[spy using outlines={3787CF, magnification=4, width={\itemwidth - 0.06cm}, height=2.4cm, connect spies,
    every spy in node/.append style={line width=0.06cm}}]
                \node [inner sep=0.0cm] {\includegraphics[width=\itemwidth]{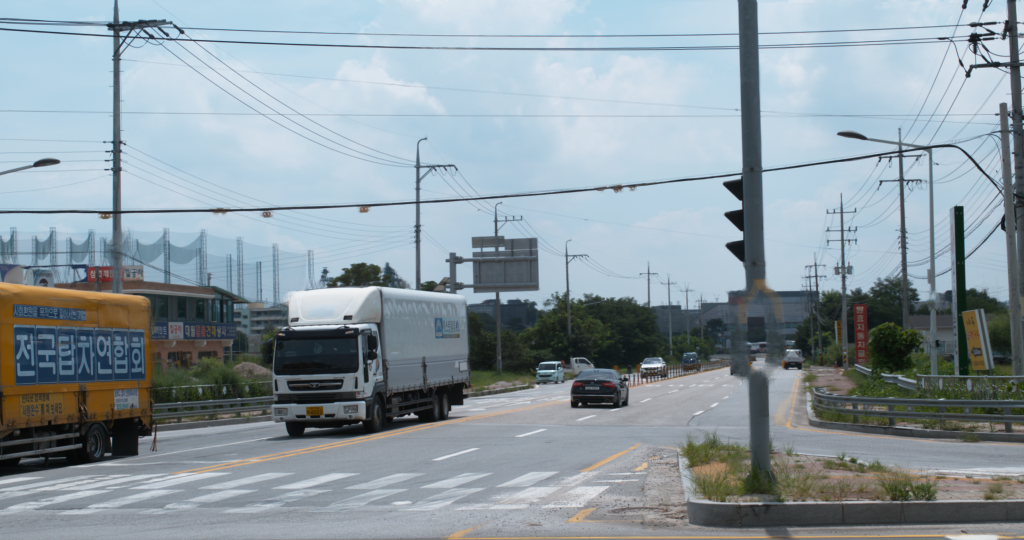}};
                \spy [every spy on node/.append style={line width=0.06cm}, spy connection path={\draw[line width=0.06cm] (tikzspyonnode) -- (tikzspyinnode);}] on (0.8,-0.27) in node at (0.0,-2.45);
            \end{tikzpicture}
        &
            \begin{tikzpicture}[spy using outlines={3787CF, magnification=4, width={\itemwidth - 0.06cm}, height=2.4cm, connect spies,
    every spy in node/.append style={line width=0.06cm}}]
                \node [inner sep=0.0cm] {\includegraphics[width=\itemwidth]{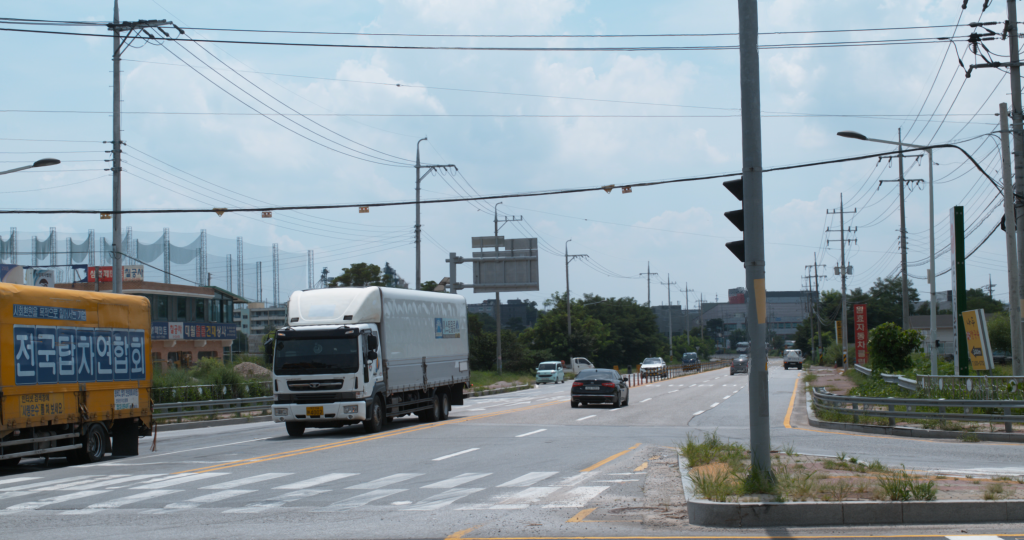}};
                \spy [every spy on node/.append style={line width=0.06cm}, spy connection path={\draw[line width=0.06cm] (tikzspyonnode) -- (tikzspyinnode);}] on (0.8,-0.27) in node at (0.0,-2.45);
            \end{tikzpicture}
        \\
            \footnotesize (a) overlaid inputs
        &
            \footnotesize (b) SoftSplat~\cite{Niklaus_CVPR_2020}
        &
            \footnotesize (c) XVFI~\cite{Sim_ICCV_2021}
        &
            \footnotesize (d) Ours
        \\
    \end{tabular}\vspace{-0.2cm}
    \caption{Qualitative comparison of our proposed approach with two representative methods on a sample from the XTEST-2K~\cite{Sim_ICCV_2021} test dataset. While these sophisticated interpolation methods are unable to handle this challenging scenario with the utility pole subject to large motion, our comparatively simple approach is able to generate a plausible result. Please consider our supplementary for more results.}\vspace{-0.4cm}
    \label{fig:qualitative}
\end{figure*}

To address such limitations, we propose a splatting-based synthesis approach. Specifically, we propose to solely rely on splatting to synthesize the output image without any subsequent refinement. As such, interpolating frames after estimating the optical flow requires only a few milliseconds and interpolating a video by a factor of $8 \times$ instead of $2 \times$ requires hardly any more compute thanks to our image formation model (Figure~\ref{fig:multime}, green). Further, our synthesis approach allows for the motion to be estimated at a lower resolution and to then upsample the estimated flow before using it to warp the input frames. This not only improves the computational efficiency, but can counterintuitively also lead to an improved interpolation quality (Figure~\ref{fig:perfplot} and Figure~\ref{fig:qualitative}).

The key to making our splatting-based synthesis approach work well is that it is carefully designed and that it is fully differentiable. Our careful design greatly improves the interpolation quality when compared to a common optical flow warping baseline ($+1.35$~dB on Vimeo-90k~\cite{Xue_IJCV_2019}), and being fully differentiable enables the underlying optical flow estimator to be fine-tuned which further improves the interpolation results ($+1.43$~dB on Vimeo-90k~\cite{Xue_IJCV_2019}). Summarizing our claims in short, we (1) introduce an image synthesis approach purely based on splatting that is especially well-suited for multi-frame interpolation, (2) show that iterative optical flow upsampling not only further improves the efficiency of our approach but can also lead to an improved quality, and (3) identify a numerical instability in softmax splatting and propose an effective solution to address it.

\section{Related Work}
\label{sec:related}

Warping-based frame interpolation has a long history. Some examples based on block-level motion estimates include overlapping block motion compensation~\cite{Choi_TCE_2000, Ha_TCE_2004}, adaptively handling overlapping blocks~\cite{Choi_OTHER_2007}, detecting and handling occlusions~\cite{Huang_TIP_2009}, considering multiple motion estimates~\cite{Jeong_TIP_2013}, and estimating a dense motion field at the interpolation instant~\cite{Dikbas_TIP_2013}. These are in contrast to motion compensation based on dense estimates which includes layered warping~\cite{Shade_OLDSIG_1998, Zitnick_TOG_2004}, occlusion reasoning for temporal interpolation~\cite{Herbst_OTHER_2009}, transition points~\cite{Mahajan_TOG_2009}, and using warping as a metric to evaluate optical flow estimates~\cite{Baker_IJCV_2011}.

Our proposed splatting-based synthesis is closely related to traditional warping techniques that leverage optical flow estimates while reasoning about occlusions~\cite{Baker_IJCV_2011, Herbst_OTHER_2009}. However, for a splatting-based synthesis approach to be used in a deep learning setting, the involved operations needs to be differentiable and easy to parallelize. This prohibits common techniques such as ordering and selecting a candidate flow in cases where multiple source pixels map to the same target~\cite{Herbst_OTHER_2009}, or iteratively filling holes~\cite{Baker_IJCV_2011}. In contrast, our proposed splatting-based synthesis technique only relies on differentiable operations that are easy to parallelize such as softmax splatting~\cite{Niklaus_CVPR_2020} and backward warping~\cite{Jaderberg_NIPS_2015}.

\begin{figure*}
    \includegraphics[]{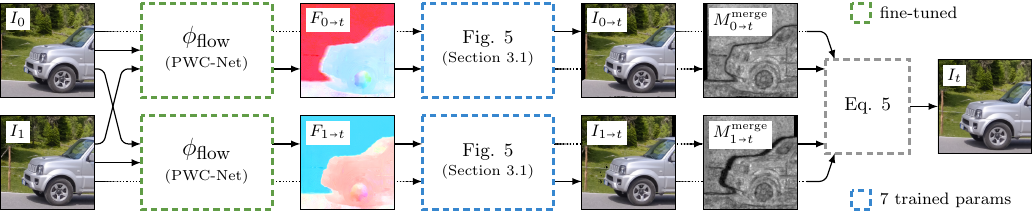}\vspace{-0.2cm}
    \caption{Overview of our proposed splatting-based synthesis for video frame interpolation. Given two frames $I_0$ and $I_1$, we estimate the inter-frame motion $F_{0 \shortto 1}$ and $F_{1 \shortto 0}$ through an off-the-shelf optical flow network. Using the flow scaled by the desired instant $t$, we then splat the input frames to time $t$ as $I_{0 \shortto t}$ and $I_{1 \shortto t}$ as outlined in Figure~\ref{fig:warpalgo} before merging them according to Equation~\ref{eqn:synthesis} to obtain $I_t$.}\vspace{-0.2cm}
    \label{fig:architecture}
\end{figure*}

\begin{figure*}
    \includegraphics[]{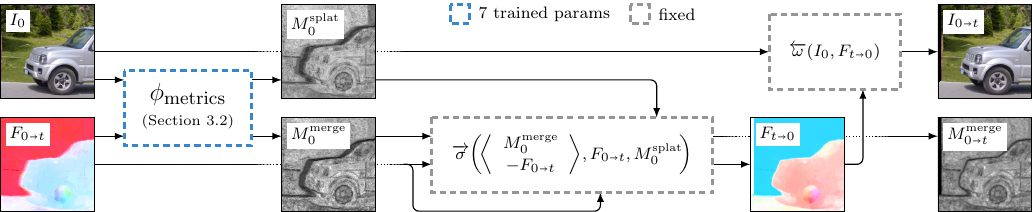}\vspace{-0.2cm}
    \caption{Given an image $I_0$ as well as an optical flow $F_{0 \shortto t}$, we not only splat the image to time $t$ as $I_{0 \shortto t}$ but also a generate a corresponding weight map $M_{0 \shortto t}^\text{merge}$ that can be used to merge multiple synthesis results. Specifically, we use softmax splatting $\protect\overrightarrow{\sigma}$~\cite{Niklaus_CVPR_2020} to splat inverse flows before employing backward warping $\protect\overleftarrow{\omega}$~\cite{Jaderberg_NIPS_2015} to reconstruct $I_{0 \shortto t}$ from $I_0$ and we directly splat a base metric $M_0^\text{merge}$ to obtain $M_{0 \shortto t}^\text{merge}$.}\vspace{-0.4cm}
    \label{fig:warpalgo}
\end{figure*}

A common category of frame interpolation approaches interpolate a frame at an arbitrary time $t$ between two input frames. We have summarized recent techniques from this category in the supplementary material since these are most closely related to our proposed approach. All of these methods have in common that they require running a neural network to infer the interpolation result at the desired instant. That is, they either use a neural network to refine warped representations of the input images, or use a neural network to infer the motion from the desired interpolation instant to the input images before accounting for it. Running such neural networks is computationally challenging though, especially at high resolutions. This is in contrast to our splatting-based synthesis where, given optical flow estimates between the input frames, synthesizing the interpolation result at any time instant requires only a few primitive operations.

% A recent notable exception is M2M ... TODO

Another category of video frame interpolation approaches take two images as input and interpolate a frame at a fixed time, typically $t = 0.5$, between the two inputs. This includes kernel-based synthesis techniques~\cite{Niklaus_CVPR_2017, Niklaus_ICCV_2017, Niklone_WACV_2021}, approaches that estimate the motion from the frame that is ought to be interpolated either implicitly~\cite{Cheng_AAAI_2020, Lee_CVPR_2020, Shi_ARXIV_2020} or explicitly~\cite{Gui_CVPR_2020, Liu_AAAI_2019, Liu_ICCV_2017, Park_ICCV_2021, Peleg_CVPR_2019, Yuan_CVPR_2019, Zhang_ECCV_2020}, methods that directly synthesize the result~\cite{Choi_AAAI_2020, Kalluri_ARXIV_2020}, and techniques that estimate the phase decomposition of the intermediate frame~\cite{Meyer_CVPR_2018}. We focus on arbitrary-time video frame interpolation.

The area of frame interpolation is much more diverse than these categories though. There is research on using multiple input frames~\cite{Chi_ECCV_2020, Liu_ARXIV_2020, Shen_CVPR_2020, Xu_NIPS_2019}, interpolating footage from event cameras~\cite{Lin_ECCV_2020, Tulyakov_CVPR_2021, Wang_OTHER_2019, Yu_ICCV_2021}, efficient model design~\cite{Choi_AAAI_2020, Choi_ICCV_2021, Ding_CVPR_2021}, test-time adaptation~\cite{Choi_CVPR_2020, Reda_ICCV_2019}, hybrid imaging systems~\cite{Paliwal_PAMI_2020}, handling quantization artifacts~\cite{Wang_CVPR_2019}, as well as joint deblurring~\cite{Shen_CVPR_2020} and super-resolution~\cite{Kim_AAAI_2020, Xiang_CVPR_2020}. Our splatting-based synthesis is orthogonal to such research directions.

\section{Splatting-based Synthesis}
\label{sec:method:interp}

Our proposed splatting-based synthesis approach for video frame interpolation is summarized in Figure~\ref{fig:architecture} and we will subsequently discuss its individual aspects. In doing so, we consider (1) how to resolve ambiguities where multiple pixels from the input image map to the same location in the target, (2) how to do the warping without introducing any unnecessary artifacts, and for video frame interpolation in particular (3) how to merge $I_0$ and $I_1$ after warping them to synthesize the desired interpolation result $I_t$ at time $t$.

\subsection{Splatting and Merging}

\begin{figure}\centering
    \setlength{\tabcolsep}{0.05cm}
    \setlength{\itemwidth}{3.2cm}
    \hspace*{-\tabcolsep}\begin{tabular}{cc}
            \begin{tikzpicture}[spy using outlines={3787CF, magnification=2.5, width=1.2cm, height={\itemwidth - 0.06cm}, connect spies,
    every spy in node/.append style={line width=0.06cm}}]
                \node [inner sep=0.0cm] {\includegraphics[width=\itemwidth]{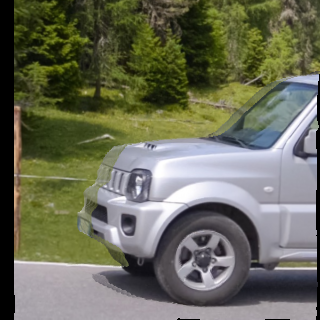}};
                \spy [every spy on node/.append style={line width=0.06cm}, spy connection path={\draw[line width=0.06cm] (tikzspyonnode) -- (tikzspyinnode);}] on (-0.7,-0.3) in node at (1.89,0.0);
            \end{tikzpicture}
        &
            \begin{tikzpicture}[spy using outlines={3787CF, magnification=2.5, width=1.2cm, height={\itemwidth - 0.06cm}, connect spies,
    every spy in node/.append style={line width=0.06cm}}]
                \node [inner sep=0.0cm] {\includegraphics[width=\itemwidth]{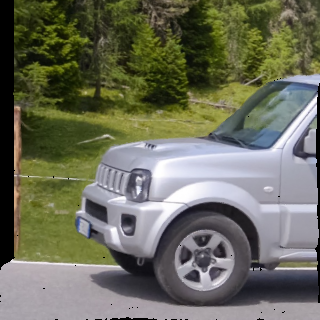}};
                \spy [every spy on node/.append style={line width=0.06cm}, spy connection path={\draw[line width=0.06cm] (tikzspyonnode) -- (tikzspyinnode);}] on (-0.7,-0.3) in node at (1.89,0.0);
            \end{tikzpicture}
        \\
            \footnotesize (a) naively splat $I_0$ to get $I_{0 \shortto t}$
        &
            \footnotesize (b) splatting weighted by $M_0^\text{splat}$
        \\
    \end{tabular}\vspace{-0.2cm}
    \caption{We use a splatting metric $M^\text{splat}$ that weights the individual pixels to resolve ambiguities where multiple pixels map to the same destination, thus properly handling occlusions.}\vspace{-0.2cm}
    \label{fig:splatmetrics}
\end{figure}

The core of our splatting-based synthesis is to warp $I_0$ and $I_1$ to the desired interpolation instant $t$ using $F_{0 \shortto t}$ and $F_{1 \shortto t}$ respectively. However, one cannot simply splat an input image as is since multiple pixels in the source image may map to the same target location as shown in Figure~\ref{fig:splatmetrics}. To address this ambiguity, we follow \cite{Niklaus_CVPR_2020} and use an auxiliary weight $M^\text{splat}$ that serves as a soft inverse z-buffer (called $Z$ in \cite{Niklaus_CVPR_2020}). We discuss how to obtain $M^\text{splat}$ in Section~\ref{sec:method:metrics}.

\begin{figure}\centering
    \setlength{\tabcolsep}{0.05cm}
    \setlength{\itemwidth}{3.2cm}
    \hspace*{-\tabcolsep}\begin{tabular}{cc}
            \begin{tikzpicture}[spy using outlines={3787CF, magnification=9, width=1.2cm, height={\itemwidth - 0.06cm}, connect spies,
    every spy in node/.append style={line width=0.06cm}}]
                \node [inner sep=0.0cm] {\includegraphics[width=\itemwidth]{warpviz-photosplat-1}};
                \spy [every spy on node/.append style={line width=0.06cm}, spy connection path={\draw[line width=0.06cm] (tikzspyonnode) -- (tikzspyinnode);}] on (0.44,-1.32) in node at (1.89,0.0);
            \end{tikzpicture}
        &
            \begin{tikzpicture}[spy using outlines={3787CF, magnification=9, width=1.2cm, height={\itemwidth - 0.06cm}, connect spies,
    every spy in node/.append style={line width=0.06cm}}]
                \node [inner sep=0.0cm] {\includegraphics[width=\itemwidth]{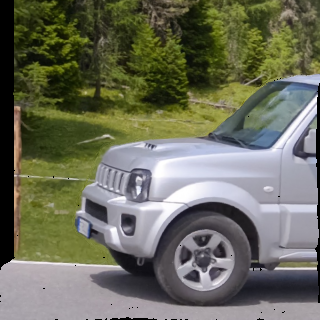}};
                \spy [every spy on node/.append style={line width=0.06cm}, spy connection path={\draw[line width=0.06cm] (tikzspyonnode) -- (tikzspyinnode);}] on (0.44,-1.32) in node at (1.89,0.0);
            \end{tikzpicture}
        \\
            \footnotesize (a) splat colors directly
        &
            \footnotesize (b) splat flows then backwarp colors
        \\
    \end{tabular}\vspace{-0.2cm}
    \caption{Directly splatting the colors of an image can lead to subtle artifacts, which is why we splat flows instead and then synthesize the output using backwards warping of the splatted flows.}\vspace{-0.2cm}
    \label{fig:splatflows}
\end{figure}

One may be tempted to directly splat $I_0$ using the optical flow $F_{0 \shortto t}$ subject to the splatting metric $M_{0}^\text{splat}$ in order to obtain $I_{0 \shortto t}$ ($I_0$ warped to time $t$). However and as shown in Figure~\ref{fig:splatflows}, this naive application of softmax splatting will lead to subtle artifacts and introduce unnecessary blurriness. Instead, we follow existing warping-based interpolation approaches and splat $F_{0 \shortto t}$ to $t$ to obtain the inverse flow $F_{t \shortto 0}$ which is then used to backward warp $I_0$ to $t$~\cite{Baker_IJCV_2011, Herbst_OTHER_2009}.

\begin{figure}\centering
    \setlength{\tabcolsep}{0.05cm}
    \setlength{\itemwidth}{3.2cm}
    \hspace*{-\tabcolsep}\begin{tabular}{cc}
            \begin{tikzpicture}[spy using outlines={3787CF, magnification=4.5, width=1.2cm, height={\itemwidth - 0.06cm}, connect spies,
    every spy in node/.append style={line width=0.06cm}}]
                \node [inner sep=0.0cm] {\includegraphics[width=\itemwidth]{warpviz-splatflows-1}};
                \spy [every spy on node/.append style={line width=0.06cm}, spy connection path={\draw[line width=0.06cm] (tikzspyonnode) -- (tikzspyinnode);}] on (-0.52,-1.2) in node at (1.89,0.0);
            \end{tikzpicture}
        &
            \begin{tikzpicture}[spy using outlines={3787CF, magnification=4.5, width=1.2cm, height={\itemwidth - 0.06cm}, connect spies,
    every spy in node/.append style={line width=0.06cm}}]
                \node [inner sep=0.0cm] {\includegraphics[width=\itemwidth]{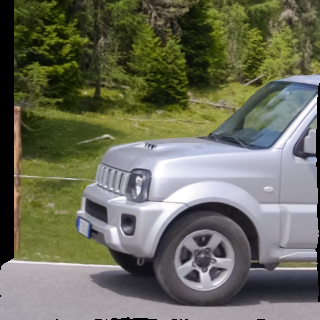}};
                \spy [every spy on node/.append style={line width=0.06cm}, spy connection path={\draw[line width=0.06cm] (tikzspyonnode) -- (tikzspyinnode);}] on (-0.52,-1.2) in node at (1.89,0.0);
            \end{tikzpicture}
        \\
            \footnotesize (a) splat flows with bilinear kernel
        &
            \footnotesize (b) splat flows with Gaussian kernel
        \\
    \end{tabular}\vspace{-0.2cm}
    \caption{Splatting is subject to holes not only due to occlusions but also due to divergent flow fields, which we address by replacing the bilinear splatting kernel with a wider Gaussian kernel.}\vspace{-0.2cm}
    \label{fig:widesplatting}
\end{figure}

Splatting naturally leads to holes in the warped result due to not only occlusions but also divergent flow fields. As shown in Figure~\ref{fig:widesplatting}, splatting with a divergent flow results in small holes even in contiguous areas. To fill these holes, we replace the default bilinear splatting kernel, which only has a footprint of $2 \times 2$, with a $4 \times 4$ Gaussian kernel. Note that such a wider kernel would lead to blurrier results when splatting colors, but it does not affect the clarity in our approach where we splat inverse flows and then backward warp the image.

After these careful considerations we are able to faithfully warp $I_0$ to $I_{0 \shortto t}$ and $I_1$ to $I_{1 \shortto t}$, but we cannot simply average these individual results to obtain the desired $I_t$ since some pixels are more reliable than others as shown in Figure~\ref{fig:mergemetrics}. As such, we introduce an auxiliary map $M^\text{merge}$ that weights the individual results before merging them to obtain $I_t$ as:
\begin{equation}\begin{aligned}\label{eqn:synthesis}
    I_t = \frac{ (1 - t) \cdot M_{0 \shortto t}^\text{merge} \cdot I_{0 \shortto t} + t \cdot M_{1 \shortto t}^\text{merge} \cdot I_{1 \shortto t} }{ (1 - t) \cdot M_{0 \shortto t}^\text{merge} + t \cdot M_{1 \shortto t}^\text{merge} }
\end{aligned}\end{equation}
where $I_{0 \shortto t}$ is $I_0$ warped to time $t$, $M_{0 \shortto t}^\text{splat}$ is $M_0^\text{splat}$ warped to time $t$, and analogous for $I_{1 \shortto t}$ and $M_{1 \shortto t}^\text{splat}$ in the opposite direction. We will subsequently describe how to obtain the involved splatting $M^\text{splat}$ and merging $M^\text{merge}$ metrics.

\subsection{Metrics for Splatting and Merging}
\label{sec:method:metrics}

\begin{figure}\centering
    \setlength{\tabcolsep}{0.05cm}
    \setlength{\itemwidth}{3.2cm}
    \hspace*{-\tabcolsep}\begin{tabular}{cc}
            \begin{tikzpicture}[spy using outlines={3787CF, magnification=2.5, width=1.2cm, height={\itemwidth - 0.06cm}, connect spies,
    every spy in node/.append style={line width=0.06cm}}]
                \node [inner sep=0.0cm] {\includegraphics[width=\itemwidth]{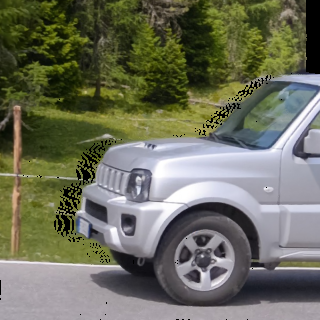}};
                \spy [every spy on node/.append style={line width=0.06cm}, spy connection path={\draw[line width=0.06cm] (tikzspyonnode) -- (tikzspyinnode);}] on (-0.8,-0.3) in node at (1.89,0.0);
            \end{tikzpicture}
        &
            \begin{tikzpicture}[spy using outlines={3787CF, magnification=2.5, width=1.2cm, height={\itemwidth - 0.06cm}, connect spies,
    every spy in node/.append style={line width=0.06cm}}]
                \node [inner sep=0.0cm] {\includegraphics[width=\itemwidth]{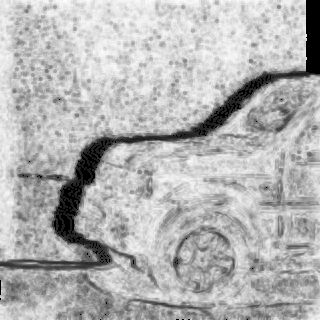}};
                \spy [every spy on node/.append style={line width=0.06cm}, spy connection path={\draw[line width=0.06cm] (tikzspyonnode) -- (tikzspyinnode);}] on (-0.8,-0.3) in node at (1.89,0.0);
            \end{tikzpicture}
        \\
            \footnotesize (a) warped image $I_{1 \shortto t}$
        &
            \footnotesize (b) corresponding $M_{1 \shortto t}^\text{merge}$
        \\
    \end{tabular}\vspace{-0.2cm}
    \caption{We use a merging metric $M^\text{merge}$ that weights the individual pixels in the warped images $I_{0 \shortto t}$ and $I_{1 \shortto t}$, which suppresses the influence of unreliable pixels when generating $I_t$.}\vspace{0.1cm}
    \label{fig:mergemetrics}
\end{figure}

Previous frame interpolation work used photometric consistency to resolve the splatting ambiguity where multiple source pixels map to the same target location~\cite{Baker_IJCV_2011}. This measure can be defined using backward warping $\overleftarrow{\omega} (\cdot)$ as:
\begin{equation}\begin{aligned}
    \psi_\text{photo} = \| I_0 - \overleftarrow{\omega} \left( I_1, F_{0 \shortto 1} \right) \|
\end{aligned}\end{equation}
However, photometric consistency is easily affected by brightness changes, as is frequently the case with moving shadows. As such, we not only consider photometric consistency but also optical flow consistency defined as:
\begin{equation}\begin{aligned}
    \psi_\text{flow} = \| F_{0 \shortto 1} + \overleftarrow{\omega} \left( F_{1 \shortto 0}, F_{0 \shortto 1} \right) \|
\end{aligned}\end{equation}
Flow consistency is given if the flow of a pixel mapped to the target maps back to the pixel in the source, which is invariant to brightness changes. Another measure we consider is flow variance, which indicates local changes in flow as:
\begin{equation}\begin{aligned}
    \psi_\text{varia} = \| \sqrt{ G(F_{0 \shortto 1} \hspace{0.0cm} ^2) - G(F_{0 \shortto 1})^2 } \|
\end{aligned}\end{equation}
where $G(\cdot)$ denotes a $3 \times 3$ Gaussian filter. Flow variance is high in areas with discontinuous flow, as is the case at motion boundaries. As shown in Figure~\ref{fig:mergemetrics}, optical flow estimates tend to be inaccurate at boundaries which makes this measure particularly useful for the $M^\text{merge}$ metric.

\begin{figure}\centering
    \setlength{\tabcolsep}{0.0cm}
    \renewcommand{\arraystretch}{1.2}
    \newcommand{\quantTit}[1]{\multicolumn{2}{c}{\scriptsize #1}}
    \newcommand{\quantSec}[1]{\scriptsize #1}
    \newcommand{\quantInd}[1]{\scriptsize #1}
    \newcommand{\quantVal}[1]{\scalebox{0.83}[1.0]{$ #1 $}}
    \newcommand{\quantFirst}[1]{\usolid{\scalebox{0.83}[1.0]{$ #1 $}}}
    \newcommand{\quantSecond}[1]{\udotted{\scalebox{0.83}[1.0]{$ #1 $}}}
    \footnotesize
    \begin{tabularx}{\columnwidth}{@{\hspace{0.1cm}} X  P{1.12cm} @{\hspace{-0.31cm}} P{1.85cm} P{1.12cm} @{\hspace{-0.31cm}} P{1.85cm}}
        \toprule
             & \quantTit{Middlebury} & \quantTit{Vimeo-90k}
        \vspace{-0.1cm}\\
             & \quantTit{Baker~\etal~\cite{Baker_IJCV_2011}} & \quantTit{Xue~\etal~\cite{Xue_IJCV_2019}}
        \\ \cmidrule(l{2pt}r{2pt}){2-3} \cmidrule(l{2pt}r{2pt}){4-5}
             & \quantSec{PSNR} \linebreak \quantInd{$\uparrow$} & {\vspace{-0.29cm} \scriptsize absolute \linebreak change} & \quantSec{PSNR} \linebreak \quantInd{$\uparrow$} & {\vspace{-0.29cm} \scriptsize absolute \linebreak change}
        \\ \midrule
Ours & \quantFirst{36.63} & \quantVal{-} & \quantFirst{35.00} & \quantVal{-}
\\
w/o flow splatting & \quantVal{36.27} & \quantVal{\text{\scalebox{1.6}[1.0]{-} } 0.36 \text{ dB}} & \quantVal{34.86} & \quantVal{\text{\scalebox{1.6}[1.0]{-} } 0.14 \text{ dB}}
\\
w/o gaussian splatting & \quantVal{36.39} & \quantVal{\text{\scalebox{1.6}[1.0]{-} } 0.24 \text{ dB}} & \quantVal{34.89} & \quantVal{\text{\scalebox{1.6}[1.0]{-} } 0.11 \text{ dB}}
\\
w/o stable splatting & \quantVal{36.48} & \quantVal{\text{\scalebox{1.6}[1.0]{-} } 0.15 \text{ dB}} & \quantVal{34.97} & \quantVal{\text{\scalebox{1.6}[1.0]{-} } 0.03 \text{ dB}}
\\
w/o using $\psi_\text{photo}$ & \quantVal{36.22} & \quantVal{\text{\scalebox{1.6}[1.0]{-} } 0.41 \text{ dB}} & \quantVal{34.99} & \quantVal{\text{\scalebox{1.6}[1.0]{-} } 0.01 \text{ dB}}
\\
w/o using $\psi_\text{flow}$ & \quantVal{36.44} & \quantVal{\text{\scalebox{1.6}[1.0]{-} } 0.19 \text{ dB}} & \quantVal{34.99} & \quantVal{\text{\scalebox{1.6}[1.0]{-} } 0.01 \text{ dB}}
\\
w/o using $\psi_\text{varia}$ & \quantVal{36.40} & \quantVal{\text{\scalebox{1.6}[1.0]{-} } 0.23 \text{ dB}} & \quantVal{34.89} & \quantVal{\text{\scalebox{1.6}[1.0]{-} } 0.11 \text{ dB}}
        \\ \bottomrule
    \end{tabularx}\vspace{-0.2cm}
    \captionof{table}{Ablative experiments to analyze the design choices of our proposed splatting-based synthesis for video frame interpolation.}\vspace{-0.2cm}
    \label{tbl:ablative}
\end{figure}

\begin{figure*}\centering
    \setlength{\tabcolsep}{0.0cm}
    \renewcommand{\arraystretch}{1.2}
    \newcommand{\quantTit}[1]{\multicolumn{2}{c}{\scriptsize #1}}
    \newcommand{\quantSec}[1]{\scriptsize #1}
    \newcommand{\quantInd}[1]{\scriptsize #1}
    \newcommand{\quantVal}[1]{\scalebox{0.83}[1.0]{$ #1 $}}
    \newcommand{\quantFirst}[1]{\usolid{\scalebox{0.83}[1.0]{$ #1 $}}}
    \newcommand{\quantSecond}[1]{\udotted{\scalebox{0.83}[1.0]{$ #1 $}}}
    \footnotesize
    \begin{tabularx}{\textwidth}{@{\hspace{0.1cm}} X  P{1.12cm} @{\hspace{-0.31cm}} P{1.85cm} P{1.12cm} @{\hspace{-0.31cm}} P{1.85cm} P{1.12cm} @{\hspace{-0.31cm}} P{1.85cm} P{1.12cm} @{\hspace{-0.31cm}} P{1.85cm} P{1.12cm} @{\hspace{-0.31cm}} P{1.85cm}}
        \toprule
             & \quantTit{Middlebury} & \quantTit{Vimeo-90k} & \quantTit{Xiph-1K} & \quantTit{Xiph-2K} & \quantTit{Xiph-4K}
        \vspace{-0.1cm}\\
             & \quantTit{Baker~\etal~\cite{Baker_IJCV_2011}} & \quantTit{Xue~\etal~\cite{Xue_IJCV_2019}} & \quantTit{(4K scaled to 1K)} & \quantTit{(4K scaled to 2K)} & \quantTit{(from xiph.org)}
        \\ \cmidrule(l{2pt}r{2pt}){2-3} \cmidrule(l{2pt}r{2pt}){4-5} \cmidrule(l{2pt}r{2pt}){6-7} \cmidrule(l{2pt}r{2pt}){8-9} \cmidrule(l{2pt}r{2pt}){10-11}
             & \quantSec{PSNR} \linebreak \quantInd{$\uparrow$} & {\vspace{-0.29cm} \scriptsize relative \linebreak change} & \quantSec{PSNR} \linebreak \quantInd{$\uparrow$} & {\vspace{-0.29cm} \scriptsize relative \linebreak change} & \quantSec{PSNR} \linebreak \quantInd{$\uparrow$} & {\vspace{-0.29cm} \scriptsize relative \linebreak change} & \quantSec{PSNR} \linebreak \quantInd{$\uparrow$} & {\vspace{-0.29cm} \scriptsize relative \linebreak change} & \quantSec{PSNR} \linebreak \quantInd{$\uparrow$} & {\vspace{-0.29cm} \scriptsize relative \linebreak change}
        \\ \midrule
fixed PWC-Net w/~\cite{Baker_IJCV_2011} warping & \quantVal{33.80} & \quantVal{-} & \quantVal{32.22} & \quantVal{-} & \quantVal{33.61} & \quantVal{-} & \quantVal{33.59} & \quantVal{-} & \quantVal{32.61} & \quantVal{-}
\\
fixed PWC-Net w/ our warping & \quantVal{34.73} & \quantVal{\text{+ } 0.93 \text{ dB}} & \quantVal{33.57} & \quantVal{\text{+ } 1.35 \text{ dB}} & \quantVal{35.03} & \quantVal{\text{+ } 1.42 \text{ dB}} & \quantVal{34.90} & \quantVal{\text{+ } 1.31 \text{ dB}} & \quantVal{33.66} & \quantVal{\text{+ } 1.05 \text{ dB}}
\\
tuned PWC-Net w/ our warping & \quantFirst{36.63} & \quantVal{\text{+ } 1.90 \text{ dB}} & \quantFirst{35.00} & \quantVal{\text{+ } 1.43 \text{ dB}} & \quantFirst{36.75} & \quantVal{\text{+ } 1.72 \text{ dB}} & \quantFirst{35.95} & \quantVal{\text{+ } 1.05 \text{ dB}} & \quantFirst{33.93} & \quantVal{\text{+ } 0.27 \text{ dB}}
        \\ \bottomrule
    \end{tabularx}\vspace{-0.2cm}
    \captionof{table}{Comparing our splatting-based synthesis to a common warping-based interpolation technique~\cite{Baker_IJCV_2011}. Not only does our approach greatly outperform this baseline, it also allows us to fine-tune the utilized PWC-Net~\cite{Sun_CVPR_2018} which further improves the interpolation results.}\vspace{-0.4cm}
    \label{tbl:middsplat}
\end{figure*}

We conclude by combining these measures and define the splatting $M^\text{splat}$ metric as (and analogous for $M^\text{merge}$):
\begin{equation}\begin{aligned}\label{eqn:metrics}
    M^\text{splat} = \frac{1}{1 \hspace{-0.05cm} + \hspace{-0.05cm} \alpha_\text{p}^\text{s} \hspace{-0.05cm} \cdot \hspace{-0.05cm} \psi_\text{photo}} \hspace{-0.05cm} + \hspace{-0.05cm} \frac{1}{1 \hspace{-0.05cm} + \hspace{-0.05cm} \alpha_\text{f}^\text{s} \hspace{-0.05cm} \cdot \hspace{-0.05cm} \psi_\text{flow}} \hspace{-0.05cm} + \hspace{-0.05cm} \frac{1}{1 \hspace{-0.05cm} + \hspace{-0.05cm} \alpha_\text{v}^\text{s} \hspace{-0.05cm} \cdot \hspace{-0.05cm} \psi_\text{varia}}
\end{aligned}\end{equation}
where $\left\langle \alpha_\text{p}^\text{s}, \alpha_\text{f}^\text{s}, \alpha_\text{v}^\text{s} \right\rangle$ are tuneable parameters. The merge metric $M^\text{merge}$ is defined analogous with $\left\langle \alpha_\text{p}^\text{m}, \alpha_\text{f}^\text{m}, \alpha_\text{v}^\text{m} \right\rangle$. We also scale $M^\text{splat}$ by an $\alpha$ as in \cite{Niklaus_CVPR_2020}, and initially set these seven parameters to $1$ while learning their values through end-to-end training. We also tried using a neural network to merge the individual measures, but have found Equation~\ref{eqn:metrics} to be faster and work better. Lastly, we also considered more complex measures such as depth~\cite{Bao_CVPR_2019} but have found these not to be beneficial due to their computational complexity.

\subsection{Ablative Experiments}

We analyze the choices we made when designing our splatting-based synthesis for frame interpolation through ablative experiments. As shown in Table~\ref{tbl:ablative}, each individual component contributes to the interpolation quality.

\subsection{Baseline Comparison}

We compare our proposed splatting-based synthesis for frame interpolation to a common warping-based interpolation technique~\cite{Baker_IJCV_2011} in Table~\ref{tbl:middsplat}, which shows that our approach greatly outperforms this common baseline. However, since our image formation model is end-to-end differentiable, we can further improve the quality of our interpolated results by fine tuning the underlying optical flow estimator. Essentially, we show how to perform the technique of~\cite{Baker_IJCV_2011} better and in a differentiable manner to enable end-to-end supervision.

\subsection{Real-time Interpolation}

Our splatting-based synthesis allows synthesizing a frame within a few milliseconds once the inter-frame motion has been estimated. We demonstrate this ability through an interactive visualization tool that is provided in the supplementary material (see Figure~\ref{fig:visualization}). This demo takes two images as well as pre-computed optical flow estimates as input and essentially implements Figure~\ref{fig:warpalgo} as well as Equation~\ref{eqn:synthesis} to synthesize the interpolated frame at the requested instant. This visualization is implemented in Javascript and it neither uses multi-threading nor any graphics acceleration. Despite this naive implementation, the demo is still able to interpolate frames in real time thanks to our image formation model.

\begin{figure}\centering
    \animategraphics[width=\columnwidth, poster=5, autoplay, palindrome, final, nomouse, method=widget]{15}{visualization-}{00000}{00010}\vspace{-0.2cm}
    \caption{An interactive demo which performs our splatting-based synthesis on the fly, please see the supplementary ``visualization.html''. This is a video that is best viewed in Adobe Reader.}\vspace{-0.4cm}
    \label{fig:visualization}
\end{figure}

\begin{figure*}
    \includegraphics[]{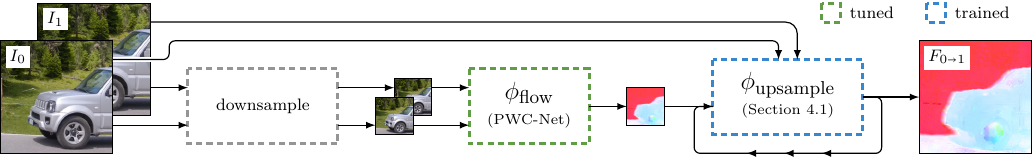}\vspace{-0.25cm}
    \caption{Overview of our iterative flow upsampling. Given two input images at a high resolution, we downsample them and then estimate the optical flow at a lower resolution. Our splatting-based synthesis requires full-resolution flow though, which is why we iteratively upsample the estimated flow guided by the input images. That is, the more we downsampled the more upsampling iterations we do.}\vspace{-0.2cm}
    \label{fig:upsaalgo}
\end{figure*}

\begin{figure*}\centering
    \setlength{\tabcolsep}{0.0cm}
    \renewcommand{\arraystretch}{1.2}
    \newcommand{\quantTit}[1]{\multicolumn{2}{c}{\scriptsize #1}}
    \newcommand{\quantTim}[1]{\multicolumn{3}{c}{\scriptsize #1}}
    \newcommand{\quantSec}[1]{\scriptsize #1}
    \newcommand{\quantInd}[1]{\scriptsize #1}
    \newcommand{\quantVal}[1]{\scalebox{0.83}[1.0]{$ #1 $}}
    \newcommand{\quantFirst}[1]{\usolid{\scalebox{0.83}[1.0]{$ #1 $}}}
    \newcommand{\quantSecond}[1]{\udotted{\scalebox{0.83}[1.0]{$ #1 $}}}
    \footnotesize
    \begin{tabularx}{\textwidth}{@{\hspace{0.1cm}} X  P{1.07cm} @{\hspace{-0.31cm}} P{1.35cm} P{1.07cm} @{\hspace{-0.31cm}} P{1.35cm} P{1.07cm} @{\hspace{-0.31cm}} P{1.35cm} P{1.07cm} @{\hspace{-0.31cm}} P{1.35cm} P{1.07cm} @{\hspace{-0.31cm}} P{1.35cm} P{1.1cm} @{\hspace{-0.31cm}} P{1.1cm} @{\hspace{-0.31cm}} P{1.1cm}}
        \toprule
             & \quantTit{Middlebury} & \quantTit{Vimeo-90k} & \quantTit{Xiph-1K} & \quantTit{Xiph-2K} & \quantTit{Xiph-4K} & \quantTim{runtime}
        \vspace{-0.1cm}\\
             & \quantTit{Baker~\etal~\cite{Baker_IJCV_2011}} & \quantTit{Xue~\etal~\cite{Xue_IJCV_2019}} & \quantTit{(4K scaled to 1K)} & \quantTit{(4K scaled to 2K)} & \quantTit{(from xiph.org)} & \quantTim{(seconds on a V100)}
        \\ \cmidrule(l{2pt}r{2pt}){2-3} \cmidrule(l{2pt}r{2pt}){4-5} \cmidrule(l{2pt}r{2pt}){6-7} \cmidrule(l{2pt}r{2pt}){8-9} \cmidrule(l{2pt}r{2pt}){10-11} \cmidrule(l{2pt}r{2pt}){12-14}
             & \quantSec{PSNR} \linebreak \quantInd{$\uparrow$} & {\vspace{-0.29cm} \scriptsize absolute \linebreak rank} & \quantSec{PSNR} \linebreak \quantInd{$\uparrow$} & {\vspace{-0.29cm} \scriptsize absolute \linebreak rank} & \quantSec{PSNR} \linebreak \quantInd{$\uparrow$} & {\vspace{-0.29cm} \scriptsize absolute \linebreak rank} & \quantSec{PSNR} \linebreak \quantInd{$\uparrow$} & {\vspace{-0.29cm} \scriptsize absolute \linebreak rank} & \quantSec{PSNR} \linebreak \quantInd{$\uparrow$} & {\vspace{-0.29cm} \scriptsize absolute \linebreak rank} & \quantSec{at 1K} \linebreak \quantInd{$\downarrow$} & \quantSec{at 2K} \linebreak \quantInd{$\downarrow$} & \quantSec{at 4K} \linebreak \quantInd{$\downarrow$}
        \\ \midrule
Ours w/o upsampling & \quantFirst{36.63} & \quantVal{1^\text{\parbox{0.15cm}{st}} \text{ of } 3} & \quantFirst{35.00} & \quantVal{1^\text{\parbox{0.15cm}{st}} \text{ of } 3} & \quantFirst{36.75} & \quantVal{1^\text{\parbox{0.15cm}{st}} \text{ of } 3} & \quantFirst{35.95} & \quantVal{1^\text{\parbox{0.15cm}{st}} \text{ of } 3} & \quantVal{33.93} & \quantVal{3^\text{\parbox{0.15cm}{rd}} \text{ of } 3} & \quantVal{0.043} & \quantVal{0.148} & \quantVal{0.589}
\\
Ours at $\nicefrac{1}{2}$ res. w/ $2 \times$ upsampling & \quantVal{34.79} & \quantVal{2^\text{\parbox{0.15cm}{nd}} \text{ of } 3} & \quantVal{33.89} & \quantVal{2^\text{\parbox{0.15cm}{nd}} \text{ of } 3} & \quantVal{35.37} & \quantVal{2^\text{\parbox{0.15cm}{nd}} \text{ of } 3} & \quantVal{35.52} & \quantVal{2^\text{\parbox{0.15cm}{nd}} \text{ of } 3} & \quantFirst{34.68} & \quantVal{1^\text{\parbox{0.15cm}{st}} \text{ of } 3} & \quantVal{0.024} & \quantVal{0.061} & \quantVal{0.226}
\\
Ours at $\nicefrac{1}{4}$ res. w/ $4 \times$ upsampling & \quantVal{33.68} & \quantVal{3^\text{\parbox{0.15cm}{rd}} \text{ of } 3} & \quantVal{32.82} & \quantVal{3^\text{\parbox{0.15cm}{rd}} \text{ of } 3} & \quantVal{34.04} & \quantVal{3^\text{\parbox{0.15cm}{rd}} \text{ of } 3} & \quantVal{34.81} & \quantVal{3^\text{\parbox{0.15cm}{rd}} \text{ of } 3} & \quantVal{34.51} & \quantVal{2^\text{\parbox{0.15cm}{nd}} \text{ of } 3} & \quantVal{0.023} & \quantVal{0.041} & \quantVal{0.137}
        \\ \bottomrule
    \end{tabularx}\vspace{-0.2cm}
    \captionof{table}{Evaluating the effect of flow upsampling on the interpolation quality and the runtime. Counterintuitively, estimating the motion on a lower resolution is not only beneficial in terms of runtime, but sometimes also quality (see $\nicefrac{1}{2}$ res. w/ $2 \times$ upsampling on Xiph-4K).}\vspace{-0.4cm}
    \label{tbl:upsampling}
\end{figure*}

\begin{figure}\centering
    \setlength{\tabcolsep}{0.0cm}
    \renewcommand{\arraystretch}{1.2}
    \newcommand{\quantTit}[1]{\multicolumn{2}{c}{\scriptsize #1}}
    \newcommand{\quantSec}[1]{\scriptsize #1}
    \newcommand{\quantInd}[1]{\scriptsize #1}
    \newcommand{\quantVal}[1]{\scalebox{0.83}[1.0]{$ #1 $}}
    \newcommand{\quantFirst}[1]{\usolid{\scalebox{0.83}[1.0]{$ #1 $}}}
    \newcommand{\quantSecond}[1]{\udotted{\scalebox{0.83}[1.0]{$ #1 $}}}
    \footnotesize
    \begin{tabularx}{\columnwidth}{@{\hspace{0.1cm}} X  P{1.12cm} @{\hspace{-0.31cm}} P{1.85cm} P{1.12cm} @{\hspace{-0.31cm}} P{1.85cm}}
        \toprule
             & \quantTit{Xiph-2K} & \quantTit{Xiph-4K}
        \vspace{-0.1cm}\\
             & \quantTit{(4K scaled to 2K)} & \quantTit{(from xiph.org)}
        \\ \cmidrule(l{2pt}r{2pt}){2-3} \cmidrule(l{2pt}r{2pt}){4-5}
             & \quantSec{PSNR} \linebreak \quantInd{$\uparrow$} & {\vspace{-0.29cm} \scriptsize relative \linebreak change} & \quantSec{PSNR} \linebreak \quantInd{$\uparrow$} & {\vspace{-0.29cm} \scriptsize relative \linebreak change}
        \\ \midrule
at $\nicefrac{1}{2}$ res. w/ bilinear up. & \quantVal{34.91} & \quantVal{-} & \quantVal{34.51} & \quantVal{-}
\\
at $\nicefrac{1}{2}$ res. w/ our up. & \quantFirst{35.52} & \quantVal{\text{+ } 0.61 \text{ dB}} & \quantFirst{34.68} & \quantVal{\text{+ } 0.17 \text{ dB}}
\\
at $\nicefrac{1}{4}$ res. w/ bilinear up. & \quantVal{32.10} & \quantVal{-} & \quantVal{33.10} & \quantVal{-}
\\
at $\nicefrac{1}{4}$ res. w/ our up. & \quantVal{34.81} & \quantVal{\text{+ } 2.71 \text{ dB}} & \quantVal{34.51} & \quantVal{\text{+ } 1.41 \text{ dB}}
        \\ \bottomrule
    \end{tabularx}\vspace{-0.2cm}
    \captionof{table}{Comparison of our iterative flow upsampling with a baseline that only uses bilinear interpolation to upsample the flow.}\vspace{-0.4cm}
    \label{tbl:bilinup}
\end{figure}

\section{Iterative Flow Upsampling}
\label{sec:method:flowup}

It is impractical to compute optical flow on a 4K video. For high-resolution inputs, we thus propose to estimate the motion at a lower resolution and then use a neural network to iteratively upsample the optical flow to the full resolution of the input (see Figure~\ref{fig:upsaalgo}). In practice, one may want to estimate the optical flow on either a 2K or a 1K resolution when given a 4K video depending on the desired performance characteristics. To support this use case, we subsequently propose an iterative optical flow upsampling approach.

\subsection{Iterative Upsampling}

We utilize a small neural network to perform iterative flow upsampling in an coarse-to-fine manner while using the high-resolution input frames as a guide. Specifically, given a flow estimate at a resolution of $\bm{x}$ as well as the two input images at a resolution of $2 \cdot \bm{x}$, the upsampling network estimates the flow at a resolution of $2 \cdot \bm{x}$ through a sequence of four convolutions with PReLU~\cite{He_ICCV_2015} activations in between. To upsample a given optical flow estimate by a factor of $4 \times$, we execute the upsampling network twice.

We have found it beneficial to not only guide the upsampling by providing the input images, but also the three measures from Section~\ref{sec:method:metrics} as they encode useful properties of the optical flow. We have otherwise kept our upsampling network deliberately simple without using spatially-varying upsampling kernels~\cite{Teed_ECCV_2020}, normalized convolution upsampling~\cite{Eldesokey_ARXIV_2021}, or self-guided upsampling~\cite{Luo_CVPR_2021}. After all, one of the reasons for estimating the optical flow at a lower resolution is improved efficiency and employing a more complex upsampling network would counteract this objective.

Another reason for estimating the optical flow a lower resolution is to mimic the inter-frame motion that the optical flow estimator was trained on during inference. In our implementation, we use PWC-Net~\cite{Sun_CVPR_2018} to estimate the optical flow and fine-tune it on input patches of size $256 \times 256$ with a relatively small inter-frame motion magnitude. This optical flow estimator is expected to perform poorly on out-of-domain high-resolution footage such as 4K inputs. But by downsampling our inputs to resemble the data on which the flow estimation network was trained, we achiever better interpolation result at high resolutions (see Table~\ref{tbl:upsampling}).

\subsection{Baseline Comparison}

We compare our proposed iterative flow upsampling to a baseline that only uses bilinear interpolation to upsample the flow in Figure~\ref{tbl:bilinup}, which shows that it is key to upsample the flow in a guided manner. Without a $\phi_\text{upsample}$ trained specifically for this task, the drop in interpolation quality, especially when estimating the motion at $\nicefrac{1}{4}$ resolution and then upsampling it by a $4 \times$, would be too severe to usefully benefit from the improved computational efficiency.

\section{Stable Softmax Splatting}
\label{sec:method:stable}

The challenge with splatting is that multiple pixels from the source image can map to the same location in the target, which creates an ambiguity that in the context of deep learning needs to be resolved differentiably. Softmax splatting is a recent solution to this problem~\cite{Niklaus_CVPR_2020}, which has already found many applications~\cite{Fan_ICCV_2021, Feng_ARXIV_2021, Holynski_CVPR_2021, Zheli_CVPR_2021, Zhao_ARXIV_2021}. However, the way softmax splatting is implemented is not numerically stable, which we subsequently outline and address.

Given an image $I_0$, an optical flow $F_{0 \shortto t}$ that maps pixels in $I_0$ to the target time $t$ and a weight map $Z_0$ to resolve ambiguities where multiple pixels from $I_0$ map to the same target location, softmax splatting $\overrightarrow{\sigma}$ is defined as:
\begin{equation}\begin{aligned}
    \overrightarrow{\sigma} \left( I_0, F_{0 \shortto t}, Z_0 \right) = \frac{ \overrightarrow{\Sigma} \left( \exp(Z_0) \cdot I_0, F_{0 \shortto t} \right) }{ \overrightarrow{\Sigma} \left( \exp(Z_0), F_{0 \shortto t} \right) }
\end{aligned}\end{equation}
where $\overrightarrow{\Sigma} (\cdot)$ is summation splatting~\cite{Niklaus_CVPR_2020} and $Z_0$ can be thought of as an importance metric that acts like a soft inverse z-buffer (a hard z-buffer is not differentiable~\cite{Nguyen_NIPS_2018}). % This is referred to as softmax splatting due to its similarity to the softmax operator.

\begin{figure*}\centering
    \setlength{\tabcolsep}{0.0cm}
    \renewcommand{\arraystretch}{1.2}
    \newcommand{\quantTit}[1]{\multicolumn{2}{c}{\scriptsize #1}}
    \newcommand{\quantSec}[1]{\scriptsize #1}
    \newcommand{\quantInd}[1]{\scriptsize #1}
    \newcommand{\quantVal}[1]{\scalebox{0.83}[1.0]{$ #1 $}}
    \newcommand{\quantFirst}[1]{\usolid{\scalebox{0.83}[1.0]{$ #1 $}}}
    \newcommand{\quantSecond}[1]{\udotted{\scalebox{0.83}[1.0]{$ #1 $}}}
    \footnotesize
    \begin{tabularx}{\textwidth}{@{\hspace{0.1cm}} X  P{1.12cm} @{\hspace{-0.31cm}} P{1.85cm} P{1.12cm} @{\hspace{-0.31cm}} P{1.85cm} P{1.12cm} @{\hspace{-0.31cm}} P{1.85cm} P{1.12cm} @{\hspace{-0.31cm}} P{1.85cm} P{1.12cm} @{\hspace{-0.31cm}} P{1.85cm}}
        \toprule
             & \quantTit{Middlebury} & \quantTit{Vimeo-90k} & \quantTit{Xiph-1K} & \quantTit{Xiph-2K} & \quantTit{Xiph-4K}
        \vspace{-0.1cm}\\
             & \quantTit{Baker~\etal~\cite{Baker_IJCV_2011}} & \quantTit{Xue~\etal~\cite{Xue_IJCV_2019}} & \quantTit{(4K scaled to 1K)} & \quantTit{(4K scaled to 2K)} & \quantTit{(from xiph.org)}
        \\ \cmidrule(l{2pt}r{2pt}){2-3} \cmidrule(l{2pt}r{2pt}){4-5} \cmidrule(l{2pt}r{2pt}){6-7} \cmidrule(l{2pt}r{2pt}){8-9} \cmidrule(l{2pt}r{2pt}){10-11}
             & \quantSec{PSNR} \linebreak \quantInd{$\uparrow$} & {\vspace{-0.29cm} \scriptsize relative \linebreak change} & \quantSec{PSNR} \linebreak \quantInd{$\uparrow$} & {\vspace{-0.29cm} \scriptsize relative \linebreak change} & \quantSec{PSNR} \linebreak \quantInd{$\uparrow$} & {\vspace{-0.29cm} \scriptsize relative \linebreak change} & \quantSec{PSNR} \linebreak \quantInd{$\uparrow$} & {\vspace{-0.29cm} \scriptsize relative \linebreak change} & \quantSec{PSNR} \linebreak \quantInd{$\uparrow$} & {\vspace{-0.29cm} \scriptsize relative \linebreak change}
        \\ \midrule
original SoftSplat~\cite{Niklaus_CVPR_2020} & \quantVal{38.42} & \quantVal{-} & \quantVal{36.10} & \quantVal{-} & \quantVal{37.96} & \quantVal{-} & \quantVal{36.62} & \quantVal{-} & \quantVal{34.20} & \quantVal{-}
\\
with our stable softmax splatting & \quantFirst{38.59} & \quantVal{\text{+ } 0.17 \text{ dB}} & \quantFirst{36.18} & \quantVal{\text{+ } 0.08 \text{ dB}} & \quantFirst{37.99} & \quantVal{\text{+ } 0.03 \text{ dB}} & \quantFirst{36.74} & \quantVal{\text{+ } 0.12 \text{ dB}} & \quantFirst{34.62} & \quantVal{\text{+ } 0.42 \text{ dB}}
        \\ \bottomrule
    \end{tabularx}\vspace{-0.2cm}
    \captionof{table}{Our stable softmax splatting formulation leads to subtle but consistent improvements when applied to the original SoftSplat~\cite{Niklaus_CVPR_2020}.}\vspace{-0.2cm}
    \label{tbl:stablesplat}
\end{figure*}

\begin{figure}\centering
    \setlength{\tabcolsep}{0.0cm}
    \renewcommand{\arraystretch}{1.2}
    \newcommand{\quantTit}[1]{\multicolumn{2}{c}{\scriptsize #1}}
    \newcommand{\quantTim}[1]{\multicolumn{3}{c}{\scriptsize #1}}
    \newcommand{\quantSec}[1]{\scriptsize #1}
    \newcommand{\quantInd}[1]{\scriptsize #1}
    \newcommand{\quantVal}[1]{\scalebox{0.83}[1.0]{$ #1 $}}
    \newcommand{\quantFirst}[1]{\usolid{\scalebox{0.83}[1.0]{$ #1 $}}}
    \newcommand{\quantSecond}[1]{\udotted{\scalebox{0.83}[1.0]{$ #1 $}}}
    \footnotesize
    \begin{tabularx}{\columnwidth}{@{\hspace{0.1cm}} X  P{1.07cm} @{\hspace{-0.31cm}} P{1.35cm} P{1.07cm} @{\hspace{-0.31cm}} P{1.35cm} P{1.07cm} @{\hspace{-0.31cm}} P{1.35cm}}
        \toprule
             & \quantTit{XTEST-1K} & \quantTit{XTEST-2K} & \quantTit{XTEST-4K}
        \vspace{-0.1cm}\\
             & \quantTit{(4K scaled to 1K)} & \quantTit{(4K scaled to 2K)} & \quantTit{Sim~\etal~\cite{Sim_ICCV_2021}}
        \\ \cmidrule(l{2pt}r{2pt}){2-3} \cmidrule(l{2pt}r{2pt}){4-5} \cmidrule(l{2pt}r{2pt}){6-7}
             & \quantSec{PSNR} \linebreak \quantInd{$\uparrow$} & {\vspace{-0.29cm} \scriptsize absolute \linebreak rank} & \quantSec{PSNR} \linebreak \quantInd{$\uparrow$} & {\vspace{-0.29cm} \scriptsize absolute \linebreak rank} & \quantSec{PSNR} \linebreak \quantInd{$\uparrow$} & {\vspace{-0.29cm} \scriptsize absolute \linebreak rank}
        \\ \midrule
SepConv~\cite{Niklaus_ICCV_2017} & \quantVal{30.35} & \quantVal{\hphantom{0}9^\text{\parbox{0.15cm}{th}} \text{ of } 16} & \quantVal{26.60} & \quantVal{11^\text{\parbox{0.15cm}{th}} \text{ of } 16} & \quantVal{24.32} & \quantVal{\hphantom{0}9^\text{\parbox{0.15cm}{th}} \text{ of } 16}
\\
CtxSyn~\cite{Niklaus_CVPR_2018} & \quantVal{31.92} & \quantVal{\hphantom{0}6^\text{\parbox{0.15cm}{th}} \text{ of } 16} & \quantVal{29.12} & \quantVal{\hphantom{0}6^\text{\parbox{0.15cm}{th}} \text{ of } 16} & \quantVal{25.46} & \quantVal{\hphantom{0}4^\text{\parbox{0.15cm}{th}} \text{ of } 16}
\\
DAIN~\cite{Bao_CVPR_2019} & \quantVal{32.51} & \quantVal{\hphantom{0}3^\text{\parbox{0.15cm}{rd}} \text{ of } 16} & \quantVal{31.49} & \quantVal{\hphantom{0}2^\text{\parbox{0.15cm}{nd}} \text{ of } 16} & \quantVal{-} & \quantVal{-}
\\
CAIN~\cite{Choi_AAAI_2020} & \quantVal{30.23} & \quantVal{11^\text{\parbox{0.15cm}{th}} \text{ of } 16} & \quantVal{26.72} & \quantVal{10^\text{\parbox{0.15cm}{th}} \text{ of } 16} & \quantVal{24.50} & \quantVal{\hphantom{0}6^\text{\parbox{0.15cm}{th}} \text{ of } 16}
\\
EDSC$_s$~\cite{Cheng_ARXIV_2020} & \quantVal{30.54} & \quantVal{\hphantom{0}8^\text{\parbox{0.15cm}{th}} \text{ of } 16} & \quantVal{26.37} & \quantVal{12^\text{\parbox{0.15cm}{th}} \text{ of } 16} & \quantVal{-} & \quantVal{-}
\\
EDSC$_m$~\cite{Cheng_ARXIV_2020} & \quantVal{29.62} & \quantVal{14^\text{\parbox{0.15cm}{th}} \text{ of } 16} & \quantVal{27.45} & \quantVal{\hphantom{0}8^\text{\parbox{0.15cm}{th}} \text{ of } 16} & \quantVal{-} & \quantVal{-}
\\
AdaCoF~\cite{Lee_CVPR_2020} & \quantVal{28.69} & \quantVal{15^\text{\parbox{0.15cm}{th}} \text{ of } 16} & \quantVal{26.20} & \quantVal{13^\text{\parbox{0.15cm}{th}} \text{ of } 16} & \quantVal{24.36} & \quantVal{\hphantom{0}7^\text{\parbox{0.15cm}{th}} \text{ of } 16}
\\
SoftSplat~\cite{Niklaus_CVPR_2020} & \quantFirst{33.42} & \quantVal{\hphantom{0}1^\text{\parbox{0.15cm}{st}} \text{ of } 16} & \quantVal{29.73} & \quantVal{\hphantom{0}5^\text{\parbox{0.15cm}{th}} \text{ of } 16} & \quantVal{25.48} & \quantVal{\hphantom{0}3^\text{\parbox{0.15cm}{rd}} \text{ of } 16}
\\
BMBC~\cite{Park_ECCV_2020} & \quantVal{30.04} & \quantVal{12^\text{\parbox{0.15cm}{th}} \text{ of } 16} & \quantVal{25.46} & \quantVal{15^\text{\parbox{0.15cm}{th}} \text{ of } 16} & \quantVal{-} & \quantVal{-}
\\
RIFE~\cite{Huang_ARXIV_2020} & \quantVal{32.32} & \quantVal{\hphantom{0}4^\text{\parbox{0.15cm}{th}} \text{ of } 16} & \quantVal{27.49} & \quantVal{\hphantom{0}7^\text{\parbox{0.15cm}{th}} \text{ of } 16} & \quantVal{24.67} & \quantVal{\hphantom{0}5^\text{\parbox{0.15cm}{th}} \text{ of } 16}
\\
SepConv++~\cite{Niklone_WACV_2021} & \quantVal{29.78} & \quantVal{13^\text{\parbox{0.15cm}{th}} \text{ of } 16} & \quantVal{26.12} & \quantVal{14^\text{\parbox{0.15cm}{th}} \text{ of } 16} & \quantVal{24.36} & \quantVal{\hphantom{0}7^\text{\parbox{0.15cm}{th}} \text{ of } 16}
\\
CDFI~\cite{Ding_CVPR_2021} & \quantVal{30.30} & \quantVal{10^\text{\parbox{0.15cm}{th}} \text{ of } 16} & \quantVal{26.89} & \quantVal{\hphantom{0}9^\text{\parbox{0.15cm}{th}} \text{ of } 16} & \quantVal{-} & \quantVal{-}
\\
XVFI~\cite{Sim_ICCV_2021} & \quantVal{31.54} & \quantVal{\hphantom{0}7^\text{\parbox{0.15cm}{th}} \text{ of } 16} & \quantVal{31.12} & \quantVal{\hphantom{0}3^\text{\parbox{0.15cm}{rd}} \text{ of } 16} & \quantVal{30.12} & \quantVal{\hphantom{0}2^\text{\parbox{0.15cm}{nd}} \text{ of } 16}
\\
XVFI$_v$~\cite{Sim_ICCV_2021} & \quantVal{26.91} & \quantVal{16^\text{\parbox{0.15cm}{th}} \text{ of } 16} & \quantVal{24.49} & \quantVal{16^\text{\parbox{0.15cm}{th}} \text{ of } 16} & \quantVal{22.83} & \quantVal{10^\text{\parbox{0.15cm}{th}} \text{ of } 16}
\\
ABME~\cite{Park_ICCV_2021} & \quantVal{32.08} & \quantVal{\hphantom{0}5^\text{\parbox{0.15cm}{th}} \text{ of } 16} & \quantVal{30.15} & \quantVal{\hphantom{0}4^\text{\parbox{0.15cm}{th}} \text{ of } 16} & \quantVal{-} & \quantVal{-}
\\
Ours & \quantVal{33.31} & \quantVal{\hphantom{0}2^\text{\parbox{0.15cm}{nd}} \text{ of } 16} & \quantFirst{32.27} & \quantVal{\hphantom{0}1^\text{\parbox{0.15cm}{st}} \text{ of } 16} & \quantFirst{31.34} & \quantVal{\hphantom{0}1^\text{\parbox{0.15cm}{st}} \text{ of } 16}
        \\ \bottomrule
    \end{tabularx}\vspace{-0.2cm}
    \captionof{table}{Evaluating the $8 \times$ interpolation capability of our approach in comparison to various other frame interpolation techniques on the XTEST~\cite{Sim_ICCV_2021} benchmark. Our approach generates better results and is an order of magnitude faster at doing so (see Figure~\ref{fig:perfplot}).}\vspace{-0.4cm}
    \label{tbl:mulframe}
\end{figure}

The softmax operator is usually not implemented as defined since it is numerically unstable, $\text{exp}(X)$ quickly exceeds 32-bit floating point when $X > 50$. Fortunately, since $\text{softmax}(X + c) = \text{softmax}(X)$ for any $c$, we can instead use $\text{softmax}(X')$ where $X' = X - \text{max}(X)$~\cite{Goodfellow_BOOK_2016}. However, one cannot directly use this trick to numerically stabilize softmax splatting. Consider a weight map $Z_0$ with one element set to $1000$ and all others $\in \left[0, 1 \right]$. Shifting the weights by $-1000$ effectively sets all but one weight to $0$ which then reduces the operation to average splatting, ignoring $Z_0$.

The weights must be shifted adaptively at the destination where multiple source pixels overlap. As such, we first warp $Z_0$ to time $t$ as $Z_{0 \shortto t}^\text{max}$ which denotes the maximum weight for each pixel in the destination. This can be efficiently computed in parallel using an atomic max. Note that this step is and need not be differentiable as it is only used to make softmax splatting numerically stable. We can then subtract $Z_{0 \shortto t}^\text{max}[\bm{p}]$ from $Z_0[\bm{q}]$ before applying the exponential function when warping from a point $\bm{q}$ to $\bm{p}$, analogous to what is typically done when implementing softmax. We thus define our numerically stable softmax splatting as:
\begin{align} \begin{split}
    \text{let } \bm{u} & = \bm{p} - \big( \bm{q} + F_{0 \shortto t}[\bm{q}] \big)
\end{split} \\[0.1cm] \begin{split}
    I_t[\bm{p}] & = \frac{ \sum_{\forall \bm{q} \in I_0} b(\bm{u}) \cdot \text{exp} \mleft( Z_0[\bm{q}] - Z_{0 \shortto t}^\text{max}[\bm{p}] \mright) \cdot I_0[\bm{q}] }{ \sum_{\forall \bm{q} \in I_0} b(\bm{u}) \cdot \text{exp} \mleft( Z_0[\bm{q}] - Z_{0 \shortto t}^\text{max}[\bm{p}] \mright) }
\end{split} \\[0.1cm] \begin{split}
    b(\bm{u}) & = \text{max} \mleft( 0, 1 - \left| \bm{u}_x \right| \mright) \cdot \text{max} \mleft( 0, 1 - \left| \bm{u}_y \right| \mright).
\end{split} \end{align}
where $b (\cdot)$ is a bilinear kernel. Next, we demonstrate the benefits of this numerically stable softmax splatting operator on the task of frame interpolation. To do so, we reimplemented SoftSplat~\cite{Niklaus_CVPR_2020} but used our numerically stable softmax splatting instead of the official implementation. As shown in Table~\ref{tbl:stablesplat}, the enhanced numerical stability of our implementation translates to subtle but consistent improvements in the interpolation quality. We expect similar improvements in other application domains such as in rolling shutter correction, video compression, video prediction, image animation, and various other synthesis tasks~\cite{Fan_ICCV_2021, Feng_ARXIV_2021, Holynski_CVPR_2021, Zheli_CVPR_2021, Zhao_ARXIV_2021}.

\begin{figure}\centering
    \hspace{-0.1cm}\includegraphics[]{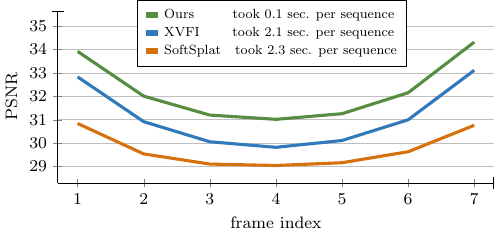}\vspace{-0.2cm}
    \caption{Evaluating the per-frame synthesis quality when performing $8 \times$ interpolation on the XTEST-2K~\cite{Sim_ICCV_2021} benchmark.}\vspace{-0.4cm}
    \label{fig:mulframe}
\end{figure}

\begin{figure*}\centering
    \setlength{\tabcolsep}{0.0cm}
    \renewcommand{\arraystretch}{1.2}
    \newcommand{\quantTit}[1]{\multicolumn{2}{c}{\scriptsize #1}}
    \newcommand{\quantTim}[1]{\multicolumn{3}{c}{\scriptsize #1}}
    \newcommand{\quantSec}[1]{\scriptsize #1}
    \newcommand{\quantInd}[1]{\scriptsize #1}
    \newcommand{\quantVal}[1]{\scalebox{0.83}[1.0]{$ #1 $}}
    \newcommand{\quantFirst}[1]{\usolid{\scalebox{0.83}[1.0]{$ #1 $}}}
    \newcommand{\quantSecond}[1]{\udotted{\scalebox{0.83}[1.0]{$ #1 $}}}
    \footnotesize
    \begin{tabularx}{\textwidth}{@{\hspace{0.1cm}} X P{1.8cm} P{1.07cm} @{\hspace{-0.31cm}} P{1.35cm} P{1.07cm} @{\hspace{-0.31cm}} P{1.35cm} P{1.07cm} @{\hspace{-0.31cm}} P{1.35cm} P{1.07cm} @{\hspace{-0.31cm}} P{1.35cm} P{1.07cm} @{\hspace{-0.31cm}} P{1.35cm} P{1.1cm} @{\hspace{-0.31cm}} P{1.1cm} @{\hspace{-0.31cm}} P{1.1cm}}
        \toprule
            & & \quantTit{Middlebury} & \quantTit{Vimeo-90k} & \quantTit{Xiph-1K} & \quantTit{Xiph-2K} & \quantTit{Xiph-4K} & \quantTim{runtime}
        \vspace{-0.1cm}\\
            & & \quantTit{Baker~\etal~\cite{Baker_IJCV_2011}} & \quantTit{Xue~\etal~\cite{Xue_IJCV_2019}} & \quantTit{(4K scaled to 1K)} & \quantTit{(4K scaled to 2K)} & \quantTit{(from xiph.org)} & \quantTim{(seconds on a V100)}
        \\ \cmidrule(l{2pt}r{2pt}){3-4} \cmidrule(l{2pt}r{2pt}){5-6} \cmidrule(l{2pt}r{2pt}){7-8} \cmidrule(l{2pt}r{2pt}){9-10} \cmidrule(l{2pt}r{2pt}){11-12} \cmidrule(l{2pt}r{2pt}){13-15}
            & {\vspace{0.04cm} \scriptsize venue} & \quantSec{PSNR} \linebreak \quantInd{$\uparrow$} & {\vspace{-0.29cm} \scriptsize absolute \linebreak rank} & \quantSec{PSNR} \linebreak \quantInd{$\uparrow$} & {\vspace{-0.29cm} \scriptsize absolute \linebreak rank} & \quantSec{PSNR} \linebreak \quantInd{$\uparrow$} & {\vspace{-0.29cm} \scriptsize absolute \linebreak rank} & \quantSec{PSNR} \linebreak \quantInd{$\uparrow$} & {\vspace{-0.29cm} \scriptsize absolute \linebreak rank} & \quantSec{PSNR} \linebreak \quantInd{$\uparrow$} & {\vspace{-0.29cm} \scriptsize absolute \linebreak rank} & \quantSec{at 1K} \linebreak \quantInd{$\downarrow$} & \quantSec{at 2K} \linebreak \quantInd{$\downarrow$} & \quantSec{at 4K} \linebreak \quantInd{$\downarrow$}
        \\ \midrule
SepConv~\cite{Niklaus_ICCV_2017} & ICCV 2017 & \quantVal{35.73} & \quantVal{12^\text{\parbox{0.15cm}{th}} \text{ of } 16} & \quantVal{33.80} & \quantVal{14^\text{\parbox{0.15cm}{th}} \text{ of } 16} & \quantVal{36.22} & \quantVal{13^\text{\parbox{0.15cm}{th}} \text{ of } 16} & \quantVal{34.77} & \quantVal{13^\text{\parbox{0.15cm}{th}} \text{ of } 16} & \quantVal{32.42} & \quantVal{\hphantom{0}7^\text{\parbox{0.15cm}{th}} \text{ of } 16} & \quantVal{0.096} & \quantVal{0.321} & \quantVal{1.245}
\\
CtxSyn~\cite{Niklaus_CVPR_2018} & CVPR 2018 & \quantVal{36.93} & \quantVal{\hphantom{0}7^\text{\parbox{0.15cm}{th}} \text{ of } 16} & \quantVal{34.39} & \quantVal{12^\text{\parbox{0.15cm}{th}} \text{ of } 16} & \quantVal{36.87} & \quantVal{\hphantom{0}5^\text{\parbox{0.15cm}{th}} \text{ of } 16} & \quantVal{35.71} & \quantVal{\hphantom{0}6^\text{\parbox{0.15cm}{th}} \text{ of } 16} & \quantVal{33.85} & \quantVal{\hphantom{0}4^\text{\parbox{0.15cm}{th}} \text{ of } 16} & \quantVal{0.111} & \quantVal{0.438} & \quantVal{1.805}
\\
DAIN~\cite{Bao_CVPR_2019} & CVPR 2019 & \quantVal{36.69} & \quantVal{10^\text{\parbox{0.15cm}{th}} \text{ of } 16} & \quantVal{34.70} & \quantVal{10^\text{\parbox{0.15cm}{th}} \text{ of } 16} & \quantVal{36.78} & \quantVal{\hphantom{0}7^\text{\parbox{0.15cm}{th}} \text{ of } 16} & \quantVal{35.93} & \quantVal{\hphantom{0}5^\text{\parbox{0.15cm}{th}} \text{ of } 16} & \quantVal{-} & \quantVal{-} & \quantVal{1.273} & \quantVal{5.679} & \quantVal{-}
\\
CAIN~\cite{Choi_AAAI_2020} & AAAI 2020 & \quantVal{35.11} & \quantVal{14^\text{\parbox{0.15cm}{th}} \text{ of } 16} & \quantVal{34.65} & \quantVal{11^\text{\parbox{0.15cm}{th}} \text{ of } 16} & \quantVal{36.21} & \quantVal{14^\text{\parbox{0.15cm}{th}} \text{ of } 16} & \quantVal{35.18} & \quantVal{\hphantom{0}9^\text{\parbox{0.15cm}{th}} \text{ of } 16} & \quantVal{32.68} & \quantVal{\hphantom{0}6^\text{\parbox{0.15cm}{th}} \text{ of } 16} & \quantVal{0.047} & \quantVal{0.158} & \quantVal{0.597}
\\
EDSC$_s$~\cite{Cheng_ARXIV_2020} & arXiv 2020 & \quantVal{36.82} & \quantVal{\hphantom{0}8^\text{\parbox{0.15cm}{th}} \text{ of } 16} & \quantVal{34.83} & \quantVal{\hphantom{0}8^\text{\parbox{0.15cm}{th}} \text{ of } 16} & \quantVal{36.73} & \quantVal{\hphantom{0}9^\text{\parbox{0.15cm}{th}} \text{ of } 16} & \quantVal{34.81} & \quantVal{12^\text{\parbox{0.15cm}{th}} \text{ of } 16} & \quantVal{-} & \quantVal{-} & \quantVal{0.961} & \quantVal{1.334} & \quantVal{-}
\\
EDSC$_m$~\cite{Cheng_ARXIV_2020} & arXiv 2020 & \quantVal{34.37} & \quantVal{15^\text{\parbox{0.15cm}{th}} \text{ of } 16} & \quantVal{33.34} & \quantVal{15^\text{\parbox{0.15cm}{th}} \text{ of } 16} & \quantVal{35.29} & \quantVal{15^\text{\parbox{0.15cm}{th}} \text{ of } 16} & \quantVal{34.62} & \quantVal{16^\text{\parbox{0.15cm}{th}} \text{ of } 16} & \quantVal{-} & \quantVal{-} & \quantVal{0.961} & \quantVal{1.334} & \quantVal{-}
\\
AdaCoF~\cite{Lee_CVPR_2020} & CVPR 2020 & \quantVal{35.72} & \quantVal{13^\text{\parbox{0.15cm}{th}} \text{ of } 16} & \quantVal{34.35} & \quantVal{13^\text{\parbox{0.15cm}{th}} \text{ of } 16} & \quantVal{36.26} & \quantVal{12^\text{\parbox{0.15cm}{th}} \text{ of } 16} & \quantVal{34.82} & \quantVal{11^\text{\parbox{0.15cm}{th}} \text{ of } 16} & \quantVal{32.12} & \quantVal{\hphantom{0}9^\text{\parbox{0.15cm}{th}} \text{ of } 16} & \quantVal{0.033} & \quantVal{0.125} & \quantVal{0.499}
\\
SoftSplat~\cite{Niklaus_CVPR_2020} & CVPR 2020 & \quantFirst{38.42} & \quantVal{\hphantom{0}1^\text{\parbox{0.15cm}{st}} \text{ of } 16} & \quantVal{36.10} & \quantVal{\hphantom{0}2^\text{\parbox{0.15cm}{nd}} \text{ of } 16} & \quantFirst{37.96} & \quantVal{\hphantom{0}1^\text{\parbox{0.15cm}{st}} \text{ of } 16} & \quantFirst{36.62} & \quantVal{\hphantom{0}1^\text{\parbox{0.15cm}{st}} \text{ of } 16} & \quantVal{34.20} & \quantVal{\hphantom{0}2^\text{\parbox{0.15cm}{nd}} \text{ of } 16} & \quantVal{0.117} & \quantVal{0.444} & \quantVal{1.768}
\\
BMBC~\cite{Park_ECCV_2020} & ECCV 2020 & \quantVal{36.79} & \quantVal{\hphantom{0}9^\text{\parbox{0.15cm}{th}} \text{ of } 16} & \quantVal{35.06} & \quantVal{\hphantom{0}6^\text{\parbox{0.15cm}{th}} \text{ of } 16} & \quantVal{36.59} & \quantVal{10^\text{\parbox{0.15cm}{th}} \text{ of } 16} & \quantVal{34.67} & \quantVal{15^\text{\parbox{0.15cm}{th}} \text{ of } 16} & \quantVal{-} & \quantVal{-} & \quantVal{1.139} & \quantVal{4.398} & \quantVal{-}
\\
RIFE~\cite{Huang_ARXIV_2020} & arXiv 2020 & \quantVal{37.30} & \quantVal{\hphantom{0}3^\text{\parbox{0.15cm}{rd}} \text{ of } 16} & \quantVal{35.61} & \quantVal{\hphantom{0}3^\text{\parbox{0.15cm}{rd}} \text{ of } 16} & \quantVal{37.38} & \quantVal{\hphantom{0}2^\text{\parbox{0.15cm}{nd}} \text{ of } 16} & \quantVal{36.16} & \quantVal{\hphantom{0}3^\text{\parbox{0.15cm}{rd}} \text{ of } 16} & \quantVal{33.47} & \quantVal{\hphantom{0}5^\text{\parbox{0.15cm}{th}} \text{ of } 16} & \quantVal{0.017} & \quantVal{0.058} & \quantVal{0.317}
\\
SepConv++~\cite{Niklone_WACV_2021} & WACV 2021 & \quantVal{37.28} & \quantVal{\hphantom{0}4^\text{\parbox{0.15cm}{th}} \text{ of } 16} & \quantVal{34.83} & \quantVal{\hphantom{0}8^\text{\parbox{0.15cm}{th}} \text{ of } 16} & \quantVal{36.83} & \quantVal{\hphantom{0}6^\text{\parbox{0.15cm}{th}} \text{ of } 16} & \quantVal{34.84} & \quantVal{10^\text{\parbox{0.15cm}{th}} \text{ of } 16} & \quantVal{32.23} & \quantVal{\hphantom{0}8^\text{\parbox{0.15cm}{th}} \text{ of } 16} & \quantVal{0.092} & \quantVal{0.364} & \quantVal{1.455}
\\
CDFI~\cite{Ding_CVPR_2021} & CVPR 2021 & \quantVal{37.14} & \quantVal{\hphantom{0}5^\text{\parbox{0.15cm}{th}} \text{ of } 16} & \quantVal{35.17} & \quantVal{\hphantom{0}4^\text{\parbox{0.15cm}{th}} \text{ of } 16} & \quantVal{37.05} & \quantVal{\hphantom{0}3^\text{\parbox{0.15cm}{rd}} \text{ of } 16} & \quantVal{35.46} & \quantVal{\hphantom{0}7^\text{\parbox{0.15cm}{th}} \text{ of } 16} & \quantVal{-} & \quantVal{-} & \quantVal{0.230} & \quantVal{0.916} & \quantVal{-}
\\
XVFI~\cite{Sim_ICCV_2021} & ICCV 2021 & \quantVal{33.27} & \quantVal{16^\text{\parbox{0.15cm}{th}} \text{ of } 16} & \quantVal{32.49} & \quantVal{16^\text{\parbox{0.15cm}{th}} \text{ of } 16} & \quantVal{34.54} & \quantVal{16^\text{\parbox{0.15cm}{th}} \text{ of } 16} & \quantVal{34.76} & \quantVal{14^\text{\parbox{0.15cm}{th}} \text{ of } 16} & \quantVal{33.99} & \quantVal{\hphantom{0}3^\text{\parbox{0.15cm}{rd}} \text{ of } 16} & \quantVal{0.114} & \quantVal{0.297} & \quantVal{0.964}
\\
XVFI$_v$~\cite{Sim_ICCV_2021} & ICCV 2021 & \quantVal{37.09} & \quantVal{\hphantom{0}6^\text{\parbox{0.15cm}{th}} \text{ of } 16} & \quantVal{35.07} & \quantVal{\hphantom{0}5^\text{\parbox{0.15cm}{th}} \text{ of } 16} & \quantVal{36.98} & \quantVal{\hphantom{0}4^\text{\parbox{0.15cm}{th}} \text{ of } 16} & \quantVal{35.19} & \quantVal{\hphantom{0}8^\text{\parbox{0.15cm}{th}} \text{ of } 16} & \quantVal{32.12} & \quantVal{\hphantom{0}9^\text{\parbox{0.15cm}{th}} \text{ of } 16} & \quantVal{0.114} & \quantVal{0.297} & \quantVal{0.964}
\\
ABME~\cite{Park_ICCV_2021} & ICCV 2021 & \quantVal{37.64} & \quantVal{\hphantom{0}2^\text{\parbox{0.15cm}{nd}} \text{ of } 16} & \quantFirst{36.18} & \quantVal{\hphantom{0}1^\text{\parbox{0.15cm}{st}} \text{ of } 16} & \quantVal{36.53} & \quantVal{11^\text{\parbox{0.15cm}{th}} \text{ of } 16} & \quantVal{36.50} & \quantVal{\hphantom{0}2^\text{\parbox{0.15cm}{nd}} \text{ of } 16} & \quantVal{-} & \quantVal{-} & \quantVal{0.336} & \quantVal{1.057} & \quantVal{-}
\\
Ours & N/A & \quantVal{36.63} & \quantVal{11^\text{\parbox{0.15cm}{th}} \text{ of } 16} & \quantVal{35.00} & \quantVal{\hphantom{0}7^\text{\parbox{0.15cm}{th}} \text{ of } 16} & \quantVal{36.75} & \quantVal{\hphantom{0}8^\text{\parbox{0.15cm}{th}} \text{ of } 16} & \quantVal{35.95} & \quantVal{\hphantom{0}4^\text{\parbox{0.15cm}{th}} \text{ of } 16} & \quantFirst{34.68} & \quantVal{\hphantom{0}1^\text{\parbox{0.15cm}{st}} \text{ of } 16} & \quantVal{0.044} & \quantVal{0.149} & \quantVal{0.226}
        \\ \bottomrule
    \end{tabularx}\vspace{-0.2cm}
    \captionof{table}{Quantitative comparison of our proposed approach with various recent frame interpolation techniques that operate on two input images. The higher the resolution the better our approach ranks, and it performs best on the Xiph-4K test where it is also the fastest.}\vspace{-0.4cm}
    \label{tbl:quantitative}
\end{figure*}

\section{Experiments}
\label{sec:experiments}

We subsequently provide additional implementation details, compare our splatting-based synthesis for frame interpolation to other approaches, and discuss its limitations.

\subsection{Implementation}

We use PWC-Net~\cite{Sun_CVPR_2018} trained on FlyingChairs~\cite{Dosovitskiy_ICCV_2015} as the basis for the underlying optical flow estimator $\phi_\text{flow}$. We fine-tune this flow estimator together with the seven parameters of the metrics extractor $\phi_\text{metrics}$ on the task of frame interpolation (Equation~\ref{eqn:synthesis})  with a Laplacian loss~\cite{Niklaus_CVPR_2018} using crops of size $256 \times 256$ from Vimeo-90k~\cite{Xue_IJCV_2019}. After convergence, we keep $\phi_\text{flow}$ and $\phi_\text{metrics}$ fixed while instead only training the iterative flow upsampling network $\phi_\text{upsample}$, again using crops from the Vimeo-90k dataset. However, this time we uniformly sample the crop width from  $\mathcal{U}(192, 448)$ and the crop height from $\mathcal{U}(192, 256)$ such that the upsampling network is supervised on various aspect ratios. During training, we run $\phi_\text{upsample}$ randomly for either one or two iterations.

\subsection{Quantitative Evaluation}

One of the benefits of our splatting-based synthesis is that once the motion has been estimated, interpolating frames only takes a few milliseconds. This makes our technique particularly useful for multi-frame interpolation, which we evaluate using the XTEST~\cite{Sim_ICCV_2021} benchmark. Since we have found the inter-frame motion in this benchmark to be rather extreme as its name suggests, we use our proposed approach with iterative $2 \times$ down/upsampling on 2K inputs while using iterative $4 \times$ down/upsampling on 4K inputs. The results of this experiment are shown in Table~\ref{tbl:mulframe} and Figure~\ref{fig:mulframe}. Aside from being highly efficient when generating multiple frames between two given ones, our approach performs particularly well on XTEST which we attribute to its favorable ability to handle large motion. Further, the per-frame analysis shows that our splatting-based synthesis is temporally consistent.

We further evaluate our approach on common benchmark datasets as done in~\cite{Niklone_WACV_2021}. For this experiment, we use our interpolation pipeline without iterative flow upsampling on inputs of up to 2K and with $2 \times$ down/upsampling for 4K inputs. As shown in Table~\ref{tbl:quantitative}, the higher the resolution the better our approach ranks and it performs best on Xiph-4K where it is also the fastest. While our approach does not yield state-of-the-art performance on low resolutions like with the Vimeo-90k test split, it is nevertheless surprising that it still outperforms both CtxSyn~\cite{Niklaus_CVPR_2018} and DAIN~\cite{Bao_CVPR_2019} on such small resolutions. After all, these methods not only splat the input images but also various feature representations before employing a synthesis network to generate the result which makes them much slower. In contrast, our synthesis is purely based on splatting without any subsequent refinement.

\subsection{Qualitative Evaluation}

Video frame interpolation results are best viewed as a motion picture, which is why we limit the qualitative evaluation in our main paper to only a single example in Figure~\ref{fig:qualitative} and kindly refer to our supplementary for more results.

\subsection{Limitations}

While generating results with our splatting-based synthesis is fast, it is wholly relying on the quality of the underlying optical flow estimate. In contrast, the refinement network that is used in related approaches that splat features before synthesizing the output using the warped features is able to account for minor inaccuracies in the estimated motion. Similarly, our splatting-based synthesis requires all the information that is necessary to interpolate the intermediate frame to be present in the input. However, this may not always be the case due to occlusions. In contrast, approaches with a refinement network can hallucinate missing content.

Furthermore, a synthesis approach like ours that solely relies on splatting will never be able to surpass an equivalent version that also utilizes a subsequent refinement network. As such, while our computational efficiency is unmatched, we consider the quantitative performance of our proposed interpolation pipeline as ``good'' but not ``state-of-the-art'' at low resolutions. The only reason we are able to claim state-of-the-art results at high resolutions is due to our iterative upsampling, but other methods could equally make use of this technique to improve their results at high resolutions.

\section{Conclusion}
\label{sec:conclusion}

In this paper, we show how to perform video frame interpolation while synthesizing the output solely through splatting. As such, synthesizing a frame only takes a few milliseconds once the inter-frame motion has been estimated, which makes our approach particularly useful for multi-frame interpolation. Furthermore, we combine this splatting-based synthesis approach with an iterative flow upsampling scheme which not only benefits the computational efficiency but also improves the interpolation quality at high resolutions.

%%%%%%%%% REFERENCES
{\small
\bibliographystyle{ieee_fullname}
\bibliography{main}

\begin{thebibliography}{10}\itemsep=-1pt

\bibitem{Baker_IJCV_2011}
Simon Baker, Daniel Scharstein, J.~P. Lewis, Stefan Roth, Michael~J. Black, and
  Richard Szeliski.
\newblock {A} {D}atabase and {E}valuation {M}ethodology for {O}ptical {F}low.
\newblock {\em {I}nternational {J}ournal of {C}omputer {V}ision}, 92(1):1--31,
  2011.

\bibitem{Bao_CVPR_2019}
Wenbo Bao, Wei-Sheng Lai, Chao Ma, Xiaoyun Zhang, Zhiyong Gao, and Ming-Hsuan
  Yang.
\newblock {D}epth-{A}ware {V}ideo {F}rame {I}nterpolation.
\newblock In {\em {IEEE} {C}onference on {C}omputer {V}ision and {P}attern
  {R}ecognition}, 2019.

\bibitem{Brooks_CVPR_2019}
Tim Brooks and Jonathan~T. Barron.
\newblock {L}earning to {S}ynthesize {M}otion {B}lur.
\newblock In {\em {IEEE} {C}onference on {C}omputer {V}ision and {P}attern
  {R}ecognition}, 2019.

\bibitem{Cheng_ARXIV_2020}
Xianhang Cheng and Zhenzhong Chen.
\newblock {M}ultiple {V}ideo {F}rame {I}nterpolation via {E}nhanced
  {D}eformable {S}eparable {C}onvolution.
\newblock {\em ar{X}iv/2006.08070}, 2020.

\bibitem{Cheng_AAAI_2020}
Xianhang Cheng and Zhenzhong Chen.
\newblock {V}ideo {F}rame {I}nterpolation via {D}eformable {S}eparable
  {C}onvolution.
\newblock In {\em {AAAI} {C}onference on {A}rtificial {I}ntelligence}, 2020.

\bibitem{Chi_ECCV_2020}
Zhixiang Chi, Rasoul~Mohammadi Nasiri, Zheng Liu, Juwei Lu, Jin Tang, and
  Konstantinos~N. Plataniotis.
\newblock {A}ll at {O}nce: {T}emporally {A}daptive {M}ulti-{F}rame
  {I}nterpolation {W}ith {A}dvanced {M}otion {M}odeling.
\newblock In {\em {E}uropean {C}onference on {C}omputer {V}ision}, 2020.

\bibitem{Choi_OTHER_2007}
Byeong-Doo Choi, Jong-Woo Han, Chang-Su Kim, and Sung-Jea Ko.
\newblock {M}otion-{C}ompensated {F}rame {I}nterpolation {U}sing {B}ilateral
  {M}otion {E}stimation and {A}daptive {O}verlapped {B}lock {M}otion
  {C}ompensation.
\newblock {\em {IEEE} {T}ransactions on {C}ircuits and {S}ystems for {V}ideo
  {T}echnology}, 17(4):407--416, 2007.

\bibitem{Choi_TCE_2000}
Byung-Tae Choi, Sung-Hee Lee, and Sung-Jea Ko.
\newblock {N}ew {F}rame {R}ate {U}p-{C}onversion {U}sing {B}i-{D}irectional
  {M}otion {E}stimation.
\newblock {\em {IEEE} {T}ransactions on {C}onsumer {E}lectronics},
  46(3):603--609, 2000.

\bibitem{Choi_CVPR_2020}
Myungsub Choi, Janghoon Choi, Sungyong Baik, Tae~Hyun Kim, and Kyoung~Mu Lee.
\newblock {S}cene-{A}daptive {V}ideo {F}rame {I}nterpolation via
  {M}eta-{L}earning.
\newblock In {\em {IEEE} {C}onference on {C}omputer {V}ision and {P}attern
  {R}ecognition}, 2020.

\bibitem{Choi_AAAI_2020}
Myungsub Choi, Heewon Kim, Bohyung Han, Ning Xu, and Kyoung~Mu Lee.
\newblock {C}hannel {A}ttention {I}s {A}ll {Y}ou {N}eed for {V}ideo {F}rame
  {I}nterpolation.
\newblock In {\em {AAAI} {C}onference on {A}rtificial {I}ntelligence}, 2020.

\bibitem{Choi_ICCV_2021}
Myungsub Choi, Suyoung Lee, Heewon Kim, and Kyoung~Mu Lee.
\newblock {M}otion-{A}ware {D}ynamic {A}rchitecture for {E}fficient {F}rame
  {I}nterpolation.
\newblock In {\em {IEEE} {I}nternational {C}onference on {C}omputer {V}ision},
  2021.

\bibitem{Dikbas_TIP_2013}
Salih Dikbas and Yucel Altunbasak.
\newblock {N}ovel {T}rue-{M}otion {E}stimation {A}lgorithm and {I}ts
  {A}pplication to {M}otion-{C}ompensated {T}emporal {F}rame {I}nterpolation.
\newblock {\em {IEEE} {T}ransactions on {I}mage {P}rocessing},
  22(8):2931--2945, 2013.

\bibitem{Ding_CVPR_2021}
Tianyu Ding, Luming Liang, Zhihui Zhu, and Ilya Zharkov.
\newblock {CDFI}: {C}ompression-{D}riven {N}etwork {D}esign for {F}rame
  {I}nterpolation.
\newblock In {\em {IEEE} {C}onference on {C}omputer {V}ision and {P}attern
  {R}ecognition}, 2021.

\bibitem{Dosovitskiy_ICCV_2015}
Alexey Dosovitskiy, Philipp Fischer, Eddy Ilg, Philip H{\"a}usser, Caner
  Hazirbas, Vladimir Golkov, Patrick van~der Smagt, Daniel Cremers, and Thomas
  Brox.
\newblock {F}low{N}et: {L}earning {O}ptical {F}low {W}ith {C}onvolutional
  {N}etworks.
\newblock In {\em {IEEE} {I}nternational {C}onference on {C}omputer {V}ision},
  2015.

\bibitem{Eldesokey_ARXIV_2021}
Abdelrahman Eldesokey and Michael Felsberg.
\newblock {N}ormalized {C}onvolution {U}psampling for {R}efined {O}ptical
  {F}low {E}stimation.
\newblock {\em ar{X}iv/2102.06979}, 2021.

\bibitem{Fan_ICCV_2021}
Bin Fan and Yuchao Dai.
\newblock {I}nverting a {R}olling {S}hutter {C}amera: {B}ring {R}olling
  {S}hutter {I}mages to {H}igh {F}ramerate {G}lobal {S}hutter {V}ideo.
\newblock In {\em {IEEE} {I}nternational {C}onference on {C}omputer {V}ision},
  2021.

\bibitem{Feng_ARXIV_2021}
Runsen Feng, Zongyu Guo, Zhizheng Zhang, and Zhibo Chen.
\newblock {V}ersatile {L}earned {V}ideo {C}ompression.
\newblock {\em ar{X}iv/2111.03386}, 2021.

\bibitem{Goodfellow_BOOK_2016}
Ian Goodfellow, Yoshua Bengio, and Aaron Courville.
\newblock {\em {D}eep {L}earning}.
\newblock MIT Press, 2016.

\bibitem{Gui_CVPR_2020}
Shurui Gui, Chaoyue Wang, Qihua Chen, and Dacheng Tao.
\newblock {F}eature{F}low: {R}obust {V}ideo {I}nterpolation via
  {S}tructure-to-{T}exture {G}eneration.
\newblock In {\em {IEEE} {C}onference on {C}omputer {V}ision and {P}attern
  {R}ecognition}, 2020.

\bibitem{Ha_TCE_2004}
Taehyeun Ha, Seongjoo Lee, and Jaeseok Kim.
\newblock {M}otion {C}ompensated {F}rame {I}nterpolation by {N}ew
  {B}lock-{B}ased {M}otion {E}stimation {A}lgorithm.
\newblock {\em {IEEE} {T}ransactions on {C}onsumer {E}lectronics},
  50(2):752--759, 2004.

\bibitem{He_ICCV_2015}
Kaiming He, Xiangyu Zhang, Shaoqing Ren, and Jian Sun.
\newblock {D}elving {D}eep {I}nto {R}ectifiers: {S}urpassing {H}uman-{L}evel
  {P}erformance on {I}mage{N}et {C}lassification.
\newblock In {\em {IEEE} {I}nternational {C}onference on {C}omputer {V}ision},
  2015.

\bibitem{Herbst_OTHER_2009}
Evan Herbst, Steve Seitz, and Simon Baker.
\newblock {O}cclusion {R}easoning for {T}emporal {I}nterpolation {U}sing
  {O}ptical {F}low.
\newblock Technical report, 2009.

\bibitem{Holynski_CVPR_2021}
Aleksander Holynski, Brian~L. Curless, Steven~M. Seitz, and Richard Szeliski.
\newblock {A}nimating {P}ictures {W}ith {E}ulerian {M}otion {F}ields.
\newblock In {\em {IEEE} {C}onference on {C}omputer {V}ision and {P}attern
  {R}ecognition}, 2021.

\bibitem{Huang_TIP_2009}
Ai-Mei Huang and Truong~Q. Nguyen.
\newblock {C}orrelation-{B}ased {M}otion {V}ector {P}rocessing {W}ith
  {A}daptive {I}nterpolation {S}cheme for {M}otion-{C}ompensated {F}rame
  {I}nterpolation.
\newblock {\em {IEEE} {T}ransactions on {I}mage {P}rocessing}, 18(4):740--752,
  2009.

\bibitem{Huang_ARXIV_2020}
Zhewei Huang, Tianyuan Zhang, Wen Heng, Boxin Shi, and Shuchang Zhou.
\newblock {RIFE}: {R}eal-{T}ime {I}ntermediate {F}low {E}stimation for {V}ideo
  {F}rame {I}nterpolation.
\newblock {\em ar{X}iv/2011.06294}, 2020.

\bibitem{Jaderberg_NIPS_2015}
Max Jaderberg, Karen Simonyan, Andrew Zisserman, and Koray Kavukcuoglu.
\newblock {S}patial {T}ransformer {N}etworks.
\newblock In {\em {A}dvances in {N}eural {I}nformation {P}rocessing {S}ystems},
  2015.

\bibitem{Jeong_TIP_2013}
Seong-Gyun Jeong, Chul Lee, and Chang-Su Kim.
\newblock {M}otion-{C}ompensated {F}rame {I}nterpolation {B}ased on
  {M}ultihypothesis {M}otion {E}stimation and {T}exture {O}ptimization.
\newblock {\em {IEEE} {T}ransactions on {I}mage {P}rocessing},
  22(11):4497--4509, 2013.

\bibitem{Kalluri_ARXIV_2020}
Tarun Kalluri, Deepak Pathak, Manmohan Chandraker, and Du Tran.
\newblock {FLAVR}: {F}low-{A}gnostic {V}ideo {R}epresentations for {F}ast
  {F}rame {I}nterpolation.
\newblock {\em ar{X}iv/2012.08512}, 2020.

\bibitem{Kim_AAAI_2020}
Soo~Ye Kim, Jihyong Oh, and Munchurl Kim.
\newblock {FISR}: {D}eep {J}oint {F}rame {I}nterpolation and
  {S}uper-{R}esolution {W}ith a {M}ulti-{S}cale {T}emporal {L}oss.
\newblock In {\em {AAAI} {C}onference on {A}rtificial {I}ntelligence}, 2020.

\bibitem{Lee_CVPR_2020}
Hyeongmin Lee, Taeoh Kim, Tae-Young Chung, Daehyun Pak, Yuseok Ban, and
  Sangyoun Lee.
\newblock {A}da{C}o{F}: {A}daptive {C}ollaboration of {F}lows for {V}ideo
  {F}rame {I}nterpolation.
\newblock In {\em {IEEE} {C}onference on {C}omputer {V}ision and {P}attern
  {R}ecognition}, 2020.

\bibitem{Sili_CVPR_2021}
Siyao Li, Shiyu Zhao, Weijiang Yu, Wenxiu Sun, Dimitris~N. Metaxas, Chen~Change
  Loy, and Ziwei Liu.
\newblock {D}eep {A}nimation {V}ideo {I}nterpolation in the {W}ild.
\newblock In {\em {IEEE} {C}onference on {C}omputer {V}ision and {P}attern
  {R}ecognition}, 2021.

\bibitem{Zheli_CVPR_2021}
Zhengqi Li, Simon Niklaus, Noah Snavely, and Oliver Wang.
\newblock {N}eural {S}cene {F}low {F}ields for {S}pace-{T}ime {V}iew
  {S}ynthesis of {D}ynamic {S}cenes.
\newblock In {\em {IEEE} {C}onference on {C}omputer {V}ision and {P}attern
  {R}ecognition}, 2021.

\bibitem{Lin_ECCV_2020}
Songnan Lin, Jiawei Zhang, Jinshan Pan, Zhe Jiang, Dongqing Zou, Yongtian Wang,
  Jing Chen, and Jimmy S.~J. Ren.
\newblock {L}earning {E}vent-{D}riven {V}ideo {D}eblurring and {I}nterpolation.
\newblock In {\em {E}uropean {C}onference on {C}omputer {V}ision}, 2020.

\bibitem{Liu_ARXIV_2020}
Yihao Liu, Liangbin Xie, Siyao Li, Wenxiu Sun, Yu Qiao, and Chao Dong.
\newblock {E}nhanced {Q}uadratic {V}ideo {I}nterpolation.
\newblock {\em ar{X}iv/2009.04642}, 2020.

\bibitem{Liu_AAAI_2019}
Yu-Lun Liu, Yi-Tung Liao, Yen-Yu Lin, and Yung-Yu Chuang.
\newblock {D}eep {V}ideo {F}rame {I}nterpolation {U}sing {C}yclic {F}rame
  {G}eneration.
\newblock In {\em {AAAI} {C}onference on {A}rtificial {I}ntelligence}, 2019.

\bibitem{Liu_ICCV_2017}
Ziwei Liu, Raymond~A. Yeh, Xiaoou Tang, Yiming Liu, and Aseem Agarwala.
\newblock {V}ideo {F}rame {S}ynthesis {U}sing {D}eep {V}oxel {F}low.
\newblock In {\em {IEEE} {I}nternational {C}onference on {C}omputer {V}ision},
  2017.

\bibitem{Luo_CVPR_2021}
Kunming Luo, Chuan Wang, Shuaicheng Liu, Haoqiang Fan, Jue Wang, and Jian Sun.
\newblock {UPF}low: {U}psampling {P}yramid for {U}nsupervised {O}ptical {F}low
  {L}earning.
\newblock In {\em {IEEE} {C}onference on {C}omputer {V}ision and {P}attern
  {R}ecognition}, 2021.

\bibitem{Mahajan_TOG_2009}
Dhruv Mahajan, Fu-Chung Huang, Wojciech Matusik, Ravi Ramamoorthi, and Peter~N.
  Belhumeur.
\newblock {M}oving {G}radients: {A} {P}ath-{B}ased {M}ethod for {P}lausible
  {I}mage {I}nterpolation.
\newblock {\em {ACM} {T}ransactions on {G}raphics}, 28(3):42:1--42:11, 2009.

\bibitem{Meyer_BMVC_2018}
Simone Meyer, Victor Cornill{\`e}re, Abdelaziz Djelouah, Christopher Schroers,
  and Markus~H. Gross.
\newblock {D}eep {V}ideo {C}olor {P}ropagation.
\newblock In {\em {B}ritish {M}achine {V}ision {C}onference}, 2018.

\bibitem{Meyer_CVPR_2018}
Simone Meyer, Abdelaziz Djelouah, Brian McWilliams, Alexander Sorkine-Hornung,
  Markus~H. Gross, and Christopher Schroers.
\newblock {P}hase{N}et for {V}ideo {F}rame {I}nterpolation.
\newblock In {\em {IEEE} {C}onference on {C}omputer {V}ision and {P}attern
  {R}ecognition}, 2018.

\bibitem{Nguyen_NIPS_2018}
Thu Nguyen-Phuoc, Chuan Li, Stephen Balaban, and Yong-Liang Yang.
\newblock {R}ender{N}et: {A} {D}eep {C}onvolutional {N}etwork for
  {D}ifferentiable {R}endering {F}rom 3{D} {S}hapes.
\newblock In {\em {A}dvances in {N}eural {I}nformation {P}rocessing {S}ystems},
  2018.

\bibitem{Niklaus_CVPR_2018}
Simon Niklaus and Feng Liu.
\newblock {C}ontext-{A}ware {S}ynthesis for {V}ideo {F}rame {I}nterpolation.
\newblock In {\em {IEEE} {C}onference on {C}omputer {V}ision and {P}attern
  {R}ecognition}, 2018.

\bibitem{Niklaus_CVPR_2020}
Simon Niklaus and Feng Liu.
\newblock {S}oftmax {S}platting for {V}ideo {F}rame {I}nterpolation.
\newblock In {\em {IEEE} {C}onference on {C}omputer {V}ision and {P}attern
  {R}ecognition}, 2020.

\bibitem{Niklaus_CVPR_2017}
Simon Niklaus, Long Mai, and Feng Liu.
\newblock {V}ideo {F}rame {I}nterpolation via {A}daptive {C}onvolution.
\newblock In {\em {IEEE} {C}onference on {C}omputer {V}ision and {P}attern
  {R}ecognition}, 2017.

\bibitem{Niklaus_ICCV_2017}
Simon Niklaus, Long Mai, and Feng Liu.
\newblock {V}ideo {F}rame {I}nterpolation via {A}daptive {S}eparable
  {C}onvolution.
\newblock In {\em {IEEE} {I}nternational {C}onference on {C}omputer {V}ision},
  2017.

\bibitem{Niklone_WACV_2021}
Simon Niklaus, Long Mai, and Oliver Wang.
\newblock {R}evisiting {A}daptive {C}onvolutions for {V}ideo {F}rame
  {I}nterpolation.
\newblock In {\em {IEEE} {W}inter {C}onference on {A}pplications of {C}omputer
  {V}ision}, 2021.

\bibitem{Nikltwo_WACV_2021}
Simon Niklaus, Xuaner~Cecilia Zhang, Jonathan~T. Barron, Neal Wadhwa, Rahul
  Garg, Feng Liu, and Tianfan Xue.
\newblock {L}earned {D}ual-{V}iew {R}eflection {R}emoval.
\newblock In {\em {IEEE} {W}inter {C}onference on {A}pplications of {C}omputer
  {V}ision}, 2021.

\bibitem{Paliwal_PAMI_2020}
Avinash Paliwal and Nima~Khademi Kalantari.
\newblock {D}eep {S}low {M}otion {V}ideo {R}econstruction {W}ith {H}ybrid
  {I}maging {S}ystem.
\newblock {\em {IEEE} {T}ransactions on {P}attern {A}nalysis and {M}achine
  {I}ntelligence}, 42(7):1557--1569, 2020.

\bibitem{Park_ECCV_2020}
Junheum Park, Keunsoo Ko, Chul Lee, and Chang-Su Kim.
\newblock {BMBC}: {B}ilateral {M}otion {E}stimation {W}ith {B}ilateral {C}ost
  {V}olume for {V}ideo {I}nterpolation.
\newblock In {\em {E}uropean {C}onference on {C}omputer {V}ision}, 2020.

\bibitem{Park_ICCV_2021}
Junheum Park, Chul Lee, and Chang-Su Kim.
\newblock {A}symmetric {B}ilateral {M}otion {E}stimation for {V}ideo {F}rame
  {I}nterpolation.
\newblock In {\em {IEEE} {I}nternational {C}onference on {C}omputer {V}ision},
  2021.

\bibitem{Peleg_CVPR_2019}
Tomer Peleg, Pablo Szekely, Doron Sabo, and Omry Sendik.
\newblock {IM}-{N}et for {H}igh {R}esolution {V}ideo {F}rame {I}nterpolation.
\newblock In {\em {IEEE} {C}onference on {C}omputer {V}ision and {P}attern
  {R}ecognition}, 2019.

\bibitem{Reda_ICCV_2019}
Fitsum~A. Reda, Deqing Sun, Aysegul Dundar, Mohammad Shoeybi, Guilin Liu,
  Kevin~J. Shih, Andrew Tao, Jan Kautz, and Bryan Catanzaro.
\newblock {U}nsupervised {V}ideo {I}nterpolation {U}sing {C}ycle {C}onsistency.
\newblock In {\em {IEEE} {I}nternational {C}onference on {C}omputer {V}ision},
  2019.

\bibitem{Shade_OLDSIG_1998}
Jonathan Shade, Steven~J. Gortler, Li wei He, and Richard Szeliski.
\newblock {L}ayered {D}epth {I}mages.
\newblock In {\em {C}onference on {C}omputer {G}raphics and {I}nteractive
  {T}echniques}, 1998.

\bibitem{Shen_CVPR_2020}
Wang Shen, Wenbo Bao, Guangtao Zhai, Li Chen, Xiongkuo Min, and Zhiyong Gao.
\newblock {B}lurry {V}ideo {F}rame {I}nterpolation.
\newblock In {\em {IEEE} {C}onference on {C}omputer {V}ision and {P}attern
  {R}ecognition}, 2020.

\bibitem{Shi_ARXIV_2020}
Zhihao Shi, Xiaohong Liu, Kangdi Shi, Linhui Dai, and Jun Chen.
\newblock {V}ideo {I}nterpolation via {G}eneralized {D}eformable {C}onvolution.
\newblock {\em ar{X}iv/2008.10680}, 2020.

\bibitem{Sim_ICCV_2021}
Hyeonjun Sim, Jihyong Oh, and Munchurl Kim.
\newblock {XVFI}: e{X}treme {V}ideo {F}rame {I}nterpolation.
\newblock In {\em {IEEE} {I}nternational {C}onference on {C}omputer {V}ision},
  2021.

\bibitem{Sun_CVPR_2018}
Deqing Sun, Xiaodong Yang, Ming-Yu Liu, and Jan Kautz.
\newblock {PWC}-{N}et: {CNN}s for {O}ptical {F}low {U}sing {P}yramid,
  {W}arping, and {C}ost {V}olume.
\newblock In {\em {IEEE} {C}onference on {C}omputer {V}ision and {P}attern
  {R}ecognition}, 2018.

\bibitem{Teed_ECCV_2020}
Zachary Teed and Jia Deng.
\newblock {RAFT}: {R}ecurrent {A}ll-{P}airs {F}ield {T}ransforms for {O}ptical
  {F}low.
\newblock In {\em {E}uropean {C}onference on {C}omputer {V}ision}, 2020.

\bibitem{Tulyakov_CVPR_2021}
Stepan Tulyakov, Daniel Gehrig, Stamatios Georgoulis, Julius Erbach, Mathias
  Gehrig, Yuanyou Li, and Davide Scaramuzza.
\newblock {T}ime {L}ens: {E}vent-{B}ased {V}ideo {F}rame {I}nterpolation.
\newblock In {\em {IEEE} {C}onference on {C}omputer {V}ision and {P}attern
  {R}ecognition}, 2021.

\bibitem{Wang_CVPR_2019}
Yang Wang, Haibin Huang, Chuan Wang, Tong He, Jue Wang, and Minh Hoai.
\newblock {GIF}2{V}ideo: {C}olor {D}equantization and {T}emporal
  {I}nterpolation of {GIF} {I}mages.
\newblock In {\em {IEEE} {C}onference on {C}omputer {V}ision and {P}attern
  {R}ecognition}, 2019.

\bibitem{Wang_OTHER_2019}
Zihao~W. Wang, Weixin Jiang, Kuan He, Boxin Shi, Aggelos~K. Katsaggelos, and
  Oliver Cossairt.
\newblock {E}vent-{D}riven {V}ideo {F}rame {S}ynthesis.
\newblock In {\em {ICCV} {W}orkshops}, 2019.

\bibitem{Wu_ECCV_2018}
Chao-Yuan Wu, Nayan Singhal, and Philipp Kr{\"a}henb{\"u}hl.
\newblock {V}ideo {C}ompression {T}hrough {I}mage {I}nterpolation.
\newblock In {\em {E}uropean {C}onference on {C}omputer {V}ision}, 2018.

\bibitem{Xiang_CVPR_2020}
Xiaoyu Xiang, Yapeng Tian, Yulun Zhang, Yun Fu, Jan~P. Allebach, and Chenliang
  Xu.
\newblock {Z}ooming {S}low-{M}o: {F}ast and {A}ccurate {O}ne-{S}tage
  {S}pace-{T}ime {V}ideo {S}uper-{R}esolution.
\newblock In {\em {IEEE} {C}onference on {C}omputer {V}ision and {P}attern
  {R}ecognition}, 2020.

\bibitem{Xu_NIPS_2019}
Xiangyu Xu, Li Si-Yao, Wenxiu Sun, Qian Yin, and Ming-Hsuan Yang.
\newblock {Q}uadratic {V}ideo {I}nterpolation.
\newblock In {\em {A}dvances in {N}eural {I}nformation {P}rocessing {S}ystems},
  2019.

\bibitem{Xue_IJCV_2019}
Tianfan Xue, Baian Chen, Jiajun Wu, Donglai Wei, and William~T. Freeman.
\newblock {V}ideo {E}nhancement {W}ith {T}ask-{O}riented {F}low.
\newblock {\em {I}nternational {J}ournal of {C}omputer {V}ision},
  127(8):1106--1125, 2019.

\bibitem{Yu_ICCV_2021}
Zhiyang Yu, Yu Zhang, Deyuan Liu, Dongqing Zou, Xijun Chen, Yebin Liu, and
  Jimmy~S. Ren.
\newblock {T}raining {W}eakly {S}upervised {V}ideo {F}rame {I}nterpolation
  {W}ith {E}vents.
\newblock In {\em {IEEE} {I}nternational {C}onference on {C}omputer {V}ision},
  2021.

\bibitem{Yuan_CVPR_2019}
Liangzhe Yuan, Yibo Chen, Hantian Liu, Tao Kong, and Jianbo Shi.
\newblock {Z}oom-in-to-{C}heck: {B}oosting {V}ideo {I}nterpolation via
  {I}nstance-{L}evel {D}iscrimination.
\newblock In {\em {IEEE} {C}onference on {C}omputer {V}ision and {P}attern
  {R}ecognition}, 2019.

\bibitem{Zhang_ECCV_2020}
Haoxian Zhang, Yang Zhao, and Ronggang Wang.
\newblock {A} {F}lexible {R}ecurrent {R}esidual {P}yramid {N}etwork for {V}ideo
  {F}rame {I}nterpolation.
\newblock In {\em {E}uropean {C}onference on {C}omputer {V}ision}, 2020.

\bibitem{Zhao_ARXIV_2021}
Lili Zhao, Zezhi Zhu, Xuhu Lin, Xuezhou Guo, Qian Yin, Wenyi Wang, and Jianwen
  Chen.
\newblock {RAI}-{N}et: {R}ange-{A}daptive {L}i{DAR} {P}oint {C}loud {F}rame
  {I}nterpolation {N}etwork.
\newblock {\em ar{X}iv/2106.00496}, 2021.

\bibitem{Zitnick_TOG_2004}
C.~Lawrence Zitnick, Sing~Bing Kang, Matthew Uyttendaele, Simon A.~J. Winder,
  and Richard Szeliski.
\newblock {H}igh-{Q}uality {V}ideo {V}iew {I}nterpolation {U}sing a {L}ayered
  {R}epresentation.
\newblock {\em {ACM} {T}ransactions on {G}raphics}, 23(3):600--608, 2004.

\end{thebibliography}
}

\end{document}